\documentclass{article}

\usepackage{PRIMEarxiv}
\usepackage[utf8]{inputenc} 
\usepackage[T1]{fontenc}    
\usepackage{hyperref}       
\usepackage{url}            
\usepackage{booktabs}       
\usepackage{textcomp}%
\usepackage{manyfoot}%
\usepackage{amsfonts}       
\usepackage{nicefrac}       
\usepackage{microtype}      
\usepackage{graphicx}
\usepackage{multirow}%
\usepackage{natbib}
\usepackage{amsmath}
\usepackage{algorithm}%
\usepackage{algorithmicx}%
\usepackage{algpseudocode}%
\usepackage{tikz}%
\usetikzlibrary{arrows.meta,angles,quotes,calc}%
\usepackage{listings}%
\usepackage{rotating}
\usepackage{makecell}

\title{SimCast-S2S: A Computationally Efficient Diffusion Model for Subseasonal Precipitation Forecasting

}

\author{
  Hiep V. Dang \\
  Department of Environmental Sciences, \\ 
  University of Virginia, Charlottesville, VA 22903, United States\\
  \texttt{zgp2ps@virginia.edu} \\
   \And
  Antonios Mamalakis \\
  School of Data Science, Department of Environmental Sciences, \\
  University of Virginia, Charlottesville, VA 22903, United States\\
  \texttt{npa4tg@virginia.edu} \\
}

\begin{document}
\maketitle

\begin{abstract}
Subseasonal-to-seasonal (S2S) precipitation forecasting has substantial financial and societal impact, yet remains challenging because of weak predictive signals, high associated uncertainty, and the computational cost of operational systems, which constrains simulation fidelity. We introduce SimCast-S2S, a generative latent-diffusion framework for probabilistic S2S precipitation forecasting that addresses three major bottlenecks in data-driven prediction. First, because S2S prediction requires uncertainty quantification rather than only deterministic point forecasts, SimCast-S2S is the first data-driven system that uses a diffusion-based generative pipeline for S2S prediction, enabling effective sampling from the underlying conditional distribution. Second, since generating large probabilistic ensembles is computationally costly in physical space, SimCast-S2S instead operates in a compact latent space learned by variational autoencoders (VAEs), enabling efficient large-ensemble generation. Third, diffusion models typically require large training datasets; SimCast-S2S overcomes this via transfer learning with low-rank adaptation (LoRA), pretraining on large ensembles of climate simulations before fine-tuning on limited reanalysis data. On reanalysis data, SimCast-S2S outperforms deep learning baselines, including convolutional neural networks and U-Net architectures. Notably, despite using only a subset of atmospheric input variables and no post-processing, bias correction, or calibration, SimCast-S2S remains competitive with, and in many aspects outperforms, state-of-the-art operational systems such as the ECMWF-S2S baseline. These results indicate that latent generative modeling combined with simulation-to-reanalysis transfer learning offers an efficient and scalable path toward data-driven probabilistic S2S precipitation forecasting.

\end{abstract}

\begingroup
\makeatletter
\renewcommand{\thefootnote}{}
\renewcommand{\@makefntext}[1]{#1}
\footnotetext{Source code for this study is available at:
\url{https://github.com/hiepdang-ml/SimCast-S2S}.}
\makeatother
\endgroup

\section{Introduction}

Subseasonal to seasonal (S2S) forecasting remains among the most important challenges in climate science, as it refers to the transition zone between weather and climate timescales ~\citep{nasesm2016next, vitart2017s2s, pegion2019subx, robertson2020s2s, white2022advances}. At short (weather) lead times, forecasts depend mainly on the initial conditions of the atmosphere, and weather prediction models can track the evolution of storms, fronts, and circulation patterns with relatively high accuracy before errors in initial conditions accumulate significantly~\citep{lorenz1969predictability, bauer2015quiet, leutbecher2008ensemble}. At seasonal and longer lead times, the focus shifts away from individual weather events and toward slowly varying boundary conditions enforced by climate variability in sea surface temperature, soil moisture, snow cover, and other land--ocean interactions that provide sources of predictability~\citep{palmer1994prospects, koster2010land, mariotti2020windows, merryfield2020current}. S2S forecasting, usually about 3 to 6 weeks ahead, lies between these two regimes~\citep{nasesm2016next, vitart2017s2s, pegion2019subx, robertson2020s2s, white2022advances}, where small errors in the initial atmospheric conditions have grown substantially, while at the same time, slower sources of variability do not yet yield stable predictive signals. Skill therefore depends on whether models can capture a mixture of fine-scale variability in atmospheric dynamics and slower Earth-system processes~\citep{zhang2005mjo, baldwin2001stratospheric, koster2010land, mariotti2020windows, merryfield2020current, richter2024sources}.

Recent progress in physical modeling and computation has made it possible to produce S2S forecasts operationally on a global scale \cite{vitart2017s2s,vitart2018s2s}. A central development in modern S2S forecasting is the use of ensemble prediction systems (EPSs)~\citep{toth1993ensemble, molteni1996ecmwf, buizza1999stochastic, gneiting2005weather, leutbecher2008ensemble, bougeault2010tigge, swinbank2016tigge, palmer2019ecmwf, vitart2017s2s}. In chaotic atmospheric systems, a single deterministic forecast is usually insufficient \cite{lorenz1969predictability,palmer2000predicting,leutbecher2008ensemble}. Thus, an EPS generates multiple forecast members by perturbing initial conditions and model physics ~\citep{toth1993ensemble, molteni1996ecmwf, buizza1999stochastic, gneiting2005weather, leutbecher2008ensemble, palmer2019ecmwf}. This approach is especially important at S2S lead times because forecast errors often grow exponentially as the atmosphere evolves before nonlinear saturation occurs~\citep{lorenz1963deterministic, lorenz1969predictability, leith1974monte, leutbecher2008ensemble}, and an EPS provides a probabilistic description of possible future states. Operational centers, such as the European Centre for Medium-Range Weather Forecasts (ECMWF), now issue S2S ensemble forecasts out to 46 days, with products often summarized as weekly anomalies relative to model climatology. The value of EPS forecasting is particularly clear for precipitation, which is one of the most difficult variables to model and predict. Precipitation is strongly affected by nonlinear processes, including convection, moisture transport, land–atmosphere feedbacks, and tropical variability \cite{robertson2020s2s}. It is widely observed that small errors in circulation, moisture transport, and humidity fields usually lead to large errors in the location and intensity of precipitation \cite{fritsch2004qpf,ebert2000precip,arakawa2004cumulus}. Forecasting centers partially address this problem with post-processing, calibration, multi-model combination, and statistical correction \cite{vannitsem2021postprocessing}, but they still face major limitations regarding systematic model biases and high computational costs~\citep{bauer2015quiet, wattmeyer2021correcting, benbouallegue2024rise}.

Statistical and machine-learning (ML) models have become an increasingly important alternative, and in many cases a complementary approach, to conventional physics-based forecasting systems \cite{rasp2020weatherbench,hwang2019ml,dang2026deepoperator,chen2024fuxis2s}. These data-driven models can exploit large archives of observations, reanalysis products, and dynamical model output to learn nonlinear relationships between atmospheric predictors and future target states. Instead of numerically solving the governing equations at every forecast step, ML models are trained to approximate the forecast mapping directly from data. This allows fast and highly parallelizable predictions during inference. Recent studies have shown that such approaches can improve S2S forecasting when they are trained on large reanalysis datasets and multi-model forecast archives \cite{hwang2019ml,chen2024fuxis2s}. Furthermore, as will be shown later, ML models pretrained on large amounts of simulated climate data can be transferred to real-world forecasting tasks through fine-tuning on observations or reanalysis products; see also \cite{weyn2021resnet,nguyen2023climax,bodnar2025aurora}. This pretraining strategy can help the model learn general atmospheric structures before adapting to the observed climate system, and it may improve forecast skill compared to models trained from scratch on limited observational data. 

Traditional EPS outputs are very computationally expensive because they require repeated model integrations with perturbed initial conditions and model physics \cite{leutbecher2008ensemble,buizza1999stochastic}. A state-of-the-art subcategory of ML methods, Generative AI, provides a framework for probabilistic S2S forecasting. Generative models aim to approximate the entire conditional distribution of future atmospheric states conditioned on relevant large-scale current climate states, allowing multiple plausible forecast samples to be drawn efficiently after training. This makes them especially useful for variables such as precipitation, where forecast uncertainty is strongly shaped by nonlinear dynamics. Recent diffusion-based forecasting systems have shown that generative models can produce realistic probabilistic forecasts and emulate large ensemble distributions at substantially lower computational cost than traditional EPSs \cite{li2024seeds,price2025gencast}.

Despite their promise, generative models also have important limitations. As noted earlier, purely data-driven models are data-thirsty, requiring large training datasets to infer atmospheric dynamics \cite{rasp2020weatherbench,dueben2018challenges}, and such datasets are often unavailable or limited, particularly beyond weather timescales. In addition, not all generative frameworks are equally stable. Generative adversarial networks (GANs), for example, have been widely used for realistic sample generation \cite{goodfellow2014gan}, but their adversarial training procedure is known to suffer from instability, mode collapse, and sensitivity to optimization settings \cite{salimans2016improved,mescheder2018convergence}. Encoder--decoder probabilistic frameworks, including models related to variational autoencoders \cite{kingma2014vae}, provide a more stable alternative and have been used in recent machine-learning S2S forecasting systems. For example, FuXi-S2S combines an encoder--decoder architecture with a perturbation module in the learned latent space, which has been shown to outperform ECMWF-S2S for selected variables, including precipitation and outgoing longwave radiation \cite{chen2024fuxis2s}, although the authors did not assess how well the model captured uncertainty. However, such models still require substantial input variables, which makes training expensive and limits their accessibility.

To address the above challenges, namely, i) the need for large forecast ensembles to characterize S2S uncertainty, ii) the large training datasets required by deep generative models, particularly under weak predictive signals and training stability issues, and iii) the high computational cost of ensemble generation, we propose SimCast-S2S, an efficient generative model based on the diffusion framework \cite{ho2020ddpm,song2021score}. SimCast-S2S uses multiple domain-specific variational autoencoders (VAEs) to compress high-dimensional physical input fields into low-dimensional Gaussian latent representations \cite{kingma2014vae}. A compact diffusion network is then trained in this latent space to map standard Gaussian noise, conditioned on past and current atmospheric states, to the future target latent representation. In the final step, a separate decoder maps the generated latent representation back to the physical space to produce the final forecast (see Methods). This design follows the general idea of latent generative modeling, where operating in a compressed representation can substantially reduce computational cost compared to generating directly in the original high-dimensional space \cite{rombach2022latentdiffusion}. 

This computational advantage is substantial in practice: on a standard A100 GPU, generating one ensemble member takes approximately 12 seconds, and different ensemble members can be produced in parallel across multiple GPUs. As a result, forecasting speed scales almost linearly with the number of GPUs and generating a e.g., 100-member forecast ensemble to better characterize forecast uncertainty becomes inexpensive compared to repeatedly running a full physics-based numerical model. Most importantly, this high degree of efficiency also makes it feasible to pre-train SimCast-S2S on a large ensemble of climate simulation output, which, we will show, addresses the need for large training datasets and substantially improves accuracy. Specifically, we train the model on 28 ensemble members of the Community Earth System Model version 2 Large Ensemble (CESM2-LE), each initialized from perturbed initial conditions \cite{rodgers2021cesm2le}. By pre-training SimCast-S2S on multiple simulation-based realizations, the model can learn a broader mapping between large-scale climate fields and future S2S precipitation variability rather than constraining training on a single and limited reanalysis dataset. The learned representation is then used for real-world forecasting by fine-tuning the model on ECMWF Reanalysis v5 (ERA5), a global atmospheric reanalysis widely used as a reference dataset for weather and climate applications \cite{hersbach2020era5}. Our results indicate that SimCast-S2S outperforms other machine-learning baselines and challenges the state-of-the-art ECMWF-S2S physics-based forecast system on the ERA5 reanalysis dataset. The improvements of SimCast-S2S are shown in terms of both deterministic forecast accuracy and probabilistic uncertainty quantification.

\begin{figure}[t]
    \centering\includegraphics[width=1\linewidth]{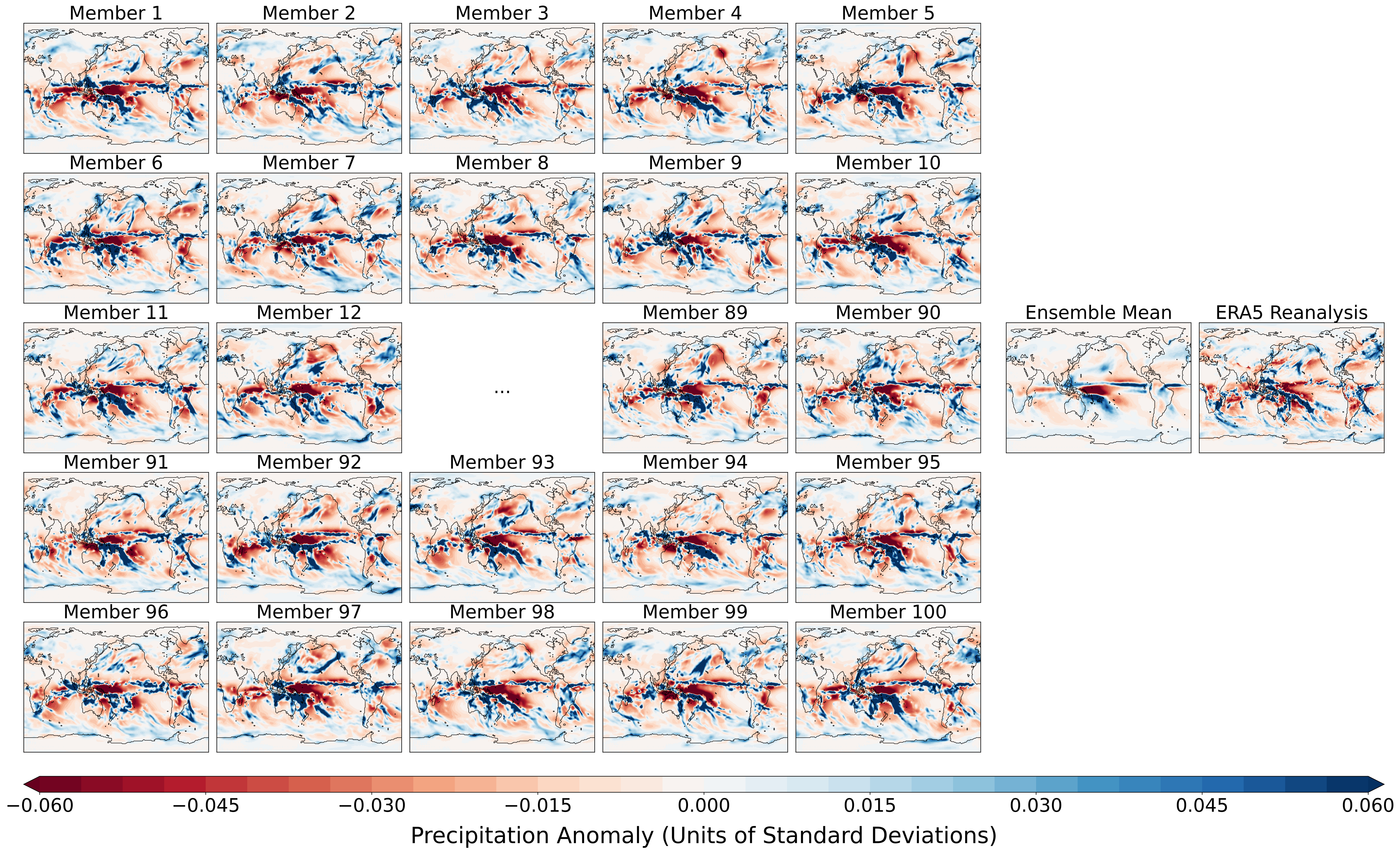}
    \caption{An ensemble of precipitation-anomaly forecasts generated by SimCast-S2S for 2-week period from February 12, 2021 to February 25, 2021 (ERA5 dataset), conditioned on the atmospheric inputs from January 01, 2021 to January 28, 2021. The ensemble mean and the corresponding ERA5 reanalysis are also provided. All fields are expressed in units of standard deviation. This example shows stronger ensemble agreement in the tropics, especially along the equatorial Pacific, and larger member-to-member variability over the extratropics. However, coherent extratropical signals remain visible over the North Pacific, the North Atlantic, and much of the Southern Hemisphere storm-track region. These areas of high consensus are also consistent with the ERA5 target.
    }
\label{fig:1}
\end{figure}

\section{Results}
\label{sec:results}

SimCast-S2S is first pretrained on 28 ensemble members from CESM2-LE using five atmospheric variables across three pressure levels 200 hPa, 500 hPa, and 850 hPa, along with five single-level variables (all considered during the last 28 days) to forecast precipitation during the 3rd and 4th week into the future. The model is trained from 1950 to 2000, validated from 2001 to 2010, and tested in years from 2011 to 2014. After pretraining, SimCast-S2S is transferred to ERA5 dataset where it is fine-tuned in the years from 1940 to 2010, validated from 2011 to 2020, and tested from 2021 to 2025 (see Section \ref{sec:method} for more details).

SimCast-S2S is computationally efficient, making it feasible to generate large forecast ensembles. A 100-member SimCast-S2S ensemble reproduces many of the large-scale precipitation anomaly structures while still retaining realistic member-to-member variability (see Fig~\ref{fig:1}). This section provides a comprehensive quantitative evaluation of the quality of the forecasts using all the test samples. Subsection~\ref{subsec:deter} compares the deterministic skill of SimCast-S2S with other ML baselines (e.g., convolutional neural networks, UNets etc.) and with the operational ECMWF-S2S system. Subsection~\ref{subsec:prob} evaluates probabilistic forecast skill across the 2021--2025 ERA5 test samples. Subsection~\ref{subsec:real} assesses the spatial realism of the generated precipitation fields. Subsection~\ref{subsec:uncertain} assesses the uncertainty reproduced by SimCast-S2S and ECMWF-S2S, and evaluates how well the forecast ranges cover the true events. Finally, subsection~\ref{subsec:eff} discusses the computational efficiency and scalability of SimCast-S2S under different hardware configurations.

\subsection{Deterministic Skill Evaluation}
\label{subsec:deter}

\begin{table}[t]
\centering
\caption{
Deterministic forecast skill measured by the mean absolute error (MAE), expressed in units of standard deviation and scaled by $10^{-2}$. We report the average MAE across all test samples and the inter-sample standard deviation. The first result column evaluates models on held-out CESM2 test data, and the remaining three result columns evaluate models on ERA5 test data under different training settings: CESM2-only training, ERA5-only training, and CESM2 pretraining followed by ERA5 fine-tuning. SimCast-S2S and ECMWF-S2S are evaluated using the ensemble mean of 100 generated members. Bold values indicate the best performance on ERA5 among different models and settings.
}
\label{tab:mae_precip}

\small
\setlength{\tabcolsep}{6pt}
\renewcommand{\arraystretch}{1.15}

\begin{tabular}{lcccc}
\multicolumn{5}{r}{{\scriptsize Unit: 10\textsuperscript{-2} std}} \\
[-1pt]
\toprule
& \multicolumn{2}{c}{\makecell{Trained on\\CESM2}}
& \multicolumn{1}{c}{\makecell{Trained on\\ERA5}}
& \multicolumn{1}{c}{\makecell{Pretrained on CESM2\\\& Finetuned on ERA5}} \\
\cmidrule(lr){2-3}
\cmidrule(lr){4-4}
\cmidrule(lr){5-5}
Model / Tested on
& CESM2
& ERA5
& ERA5
& ERA5 \\
\midrule

\multicolumn{5}{l}{\textit{Proposed generative model}} \\
SimCast-S2S ($\eta = 0.0$)
& $1.25 \pm 0.06$
& $1.32 \pm 0.06$
& $1.37 \pm 0.06$
& $\mathbf{1.30 \pm 0.06}$ \\
SimCast-S2S ($\eta = 0.5$)
& $1.25 \pm 0.06$
& $1.32 \pm 0.06$
& $1.37 \pm 0.06$
& $\mathbf{1.30 \pm 0.06}$ \\
SimCast-S2S ($\eta = 1.0$)
& $1.25 \pm 0.06$
& $1.32 \pm 0.07$
& $1.38 \pm 0.07$
& $\mathbf{1.30 \pm 0.06}$ \\

\midrule
\multicolumn{5}{l}{\textit{Neural-network baselines}} \\
CNN-Small
& $1.43 \pm 0.08$
& $1.42 \pm 0.06$
& $1.34 \pm 0.07$
& $1.38 \pm 0.06$ \\
CNN-Medium
& $1.31 \pm 0.07$
& $1.32 \pm 0.06$
& $1.46 \pm 0.06$
& $1.33 \pm 0.06$ \\
CNN-Large
& $1.51 \pm 0.09$
& $1.47 \pm 0.08$
& $1.47 \pm 0.07$
& $1.50 \pm 0.08$ \\
UNet-Small
& $1.35 \pm 0.08$
& $1.35 \pm 0.06$
& $1.36 \pm 0.07$
& $1.35 \pm 0.07$ \\
UNet-Medium
& $1.32 \pm 0.08$
& $1.32 \pm 0.06$
& $1.36 \pm 0.08$
& $1.33 \pm 0.06$ \\
UNet-Large
& $1.31 \pm 0.07$
& $1.32 \pm 0.06$
& $1.35 \pm 0.07$
& $1.33 \pm 0.06$ \\

\specialrule{0.8pt}{3pt}{3pt}
\multicolumn{5}{l}{\textit{Operational baseline}} \\
ECMWF-S2S
&
&
&
& $1.32 \pm 0.06$ \\

\bottomrule
\end{tabular}
\end{table}

Although the atmosphere is highly chaotic and a deterministic forecast cannot represent the full range of possible future states, deterministic skill remains an essential first-order measure of forecast quality, as it assesses the conditional mean of the underlying distribution of future states. In an ensemble forecasting system, the ensemble mean summarizes the predictable component shared across members by averaging out member-specific variability. If the individual ensemble members are centered around physically meaningful future states, the ensemble mean should provide an accurate estimate of the expected precipitation anomalous field. Evaluating deterministic skill tests whether SimCast-S2S captures the dominant predictable signal and is also necessary for fair comparison with popular Deep Learning (DL) baselines, which often produce deterministic predictions. We have evaluated the deterministic skill of SimCast-S2S on the ERA5 test samples using its ensemble-mean prediction, and compared it against the ensemble-mean prediction of ECMWF-S2S and deterministic predictions from different variants of Convolutional Neural Networks (CNNs) \cite{lecun1998cnn} and UNets \cite{ronneberger2015unet}. Each CNN and UNet variant corresponds to a specific model size---small, medium, or large---determined by the number of layers and hidden features. Architectural details are provided in Section~\ref{sec:method}.

\begin{figure}[t]
    \centering
    \includegraphics[width=0.8\linewidth]{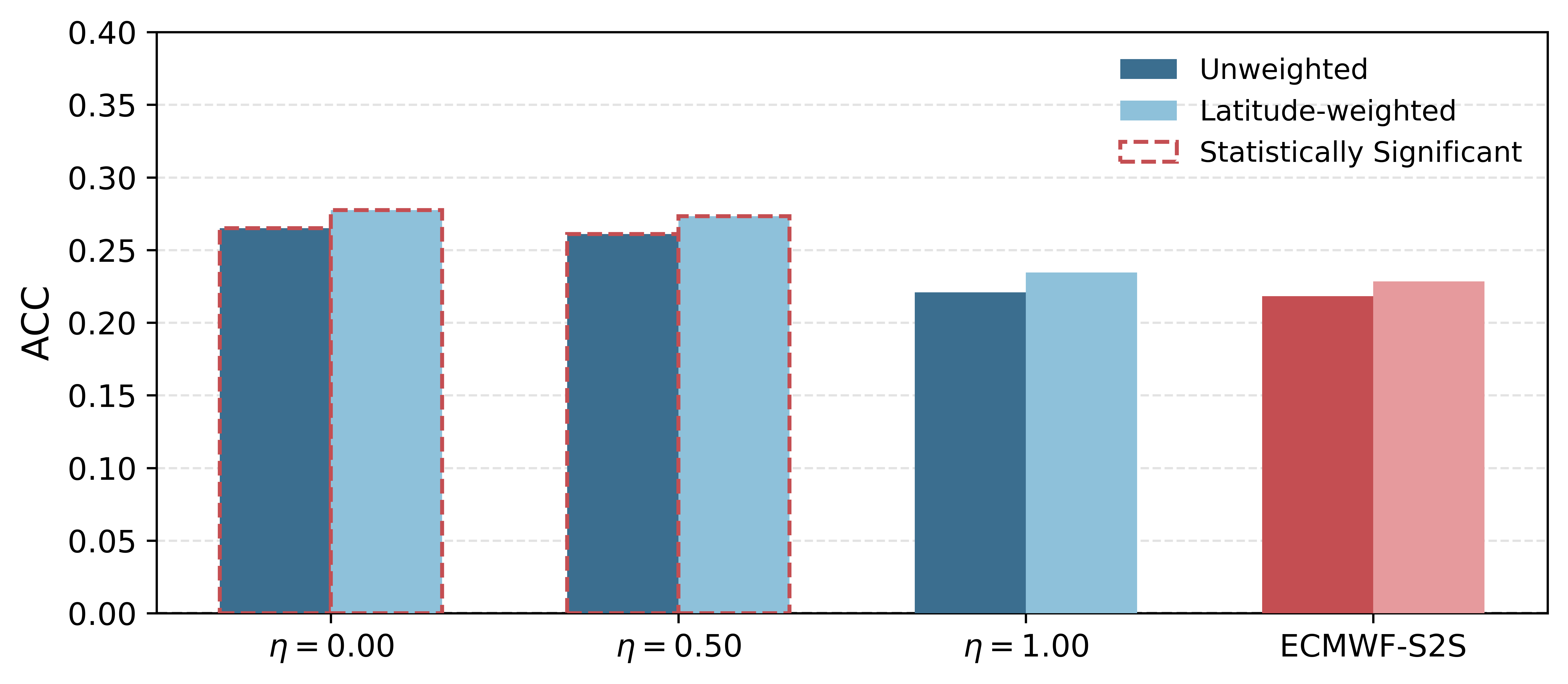}
    \caption{
    Validation anomaly correlation coefficient (ACC) for SimCast-S2S under different sampling parameters $\eta$ and for ECMWF-S2S. ACC is calculated using the ensemble mean forecast. Dark bars show unweighted ACC, and light bars show latitude-weighted ACC. Red outlines indicate cases where SimCast-S2S has significantly higher ACC than ECMWF-S2S under a one-sided bootstrap test at the $2.5\%$ significance level.
    }
\label{fig:acc}
\end{figure}

On the CESM2 test set, SimCast-S2S achieves the lowest mean absolute error (MAE) among all evaluated models and training settings (see the first column in Table~\ref{tab:mae_precip}). All three sampling settings obtain $(1.25 \pm 0.06)\times10^{-2}$ std, while the best CNN and UNet baselines reach only $(1.31 \pm 0.07)\times10^{-2}$ std. The baseline results also show that increasing model size does not guarantee better performance. CNN-Large performs worse than CNN-Medium, indicating possible overfitting. Similarly, UNet-Large has substantially more parameters than UNet-Medium, but it produces nearly the same error, which suggests that the UNet architecture gains little from further scaling.
SimCast-S2S also exhibits competitive simulation-to-reanalysis transfer without ERA5 fine-tuning. When trained only on CESM2 and evaluated directly on ERA5, SimCast-S2S achieves an MAE of $1.32\times10^{-2}$ std, lower than most CNN and UNet baselines and comparable to the best-performing UNet variants (see the second column in Table~\ref{tab:mae_precip}). This result indicates that CESM2 training alone learns precipitation features that partially transfer to ERA5. However, the remaining error also shows a clear simulation-to-reanalysis gap.
This gap (although marginal) may be partially attributable to imperfect/biased representations of the S2S relevant processes in CESM2 simulations that have led to a slightly compromised learning that is revealed when models are tested on real-world data. Regarding different sizes of the baseline DL models, results mirror the behavior of when tested on the CESM2 test set. SimCast-S2S performs worst when trained only on ERA5 (see the third column in \ref{tab:mae_precip}), with MAE increasing to $1.37$--$1.38 \times 10^{-2}$ std, a large degradation especially when compared to $1.32 \times 10^{-2}$ std in CESM2 training. This result indicates that the diffusion-based SimCast-S2S requires more training data, and training on ERA5 alone was not sufficient. Large simulated ensembles are therefore important for training the generative mapping before adapting the model to reanalysis data~\cite{rasp2021data,ham2019deep,gonzalezabad2026pretraining,unal2023climate,martin2025long}.
Pretraining on CESM2 followed by ERA5 fine-tuning yields the best performance on the ERA5 test set (see the last column in Table~\ref{tab:mae_precip}). Under this setting, all three SimCast-S2S variants achieve $(1.30 \pm 0.06)\times10^{-2}$ std, which is the lowest MAE on ERA5 in the table. SimCast-S2S outperforms all CNN and UNet baselines and also improves over the operational ECMWF-S2S benchmark, which obtains $(1.32 \pm 0.06)\times10^{-2}$ std on the same ERA5 test samples. It is clear that CESM2 pretraining exposes the model to a larger set of physically plausible atmospheric evolutions and ERA5 fine-tuning adapts this learned representation to the observation-constrained atmosphere. Transfer learning is the key strategy that allows SimCast-S2S to overcome a central limitation of generative models, namely their dependence on large training datasets. The simulation-to-reanalysis transfer strategy, rather than the architecture alone, enables SimCast-S2S to outperform the operational baseline. More details on the transfer learning method are discussed in Section~\ref{subsec:lora}.

The deterministic results also show that the choice of sampling parameter $\eta$, which controls the trade-off between diversity and stability (see Subsection \ref{subsec:simcasts2s} in Methods), has little effect on the ensemble-mean MAE. Across all training settings, the three SimCast-S2S variants produce nearly identical deterministic errors. This is somewhat expected because MAE is evaluated using the ensemble mean, which averages out member-specific sampling variability. However, different $\eta$ values result in different anomaly correlation coefficients (ACC). SimCast-S2S achieves substantially higher mean ACC than ECMWF-S2S for $\eta=0$ and $\eta=0.5$, whereas its performance for $\eta=1.0$ is closer to that of ECMWF-S2S over the 2021--2025 validation period (see Fig.~\ref{fig:acc}). Given the similar MAE across sampling settings, the higher ACC for $\eta=0$ and $\eta=0.5$ indicates better spatial correspondence between the predicted and the true precipitation patterns rather than simply smaller point-wise error magnitudes. As described in Section~\ref{sec:method}, $\eta$ controls the sampling behavior of SimCast-S2S and is not relevant to training. We therefore treat $\eta$ as an untrainable hyperparameter and select it using validation ACC reported in Fig~\ref{fig:acc}. Among the three experimented values, $\eta=0$ achieves the highest ACC. We use $\eta=0$ as the default sampling setting for SimCast-S2S, and all remaining analyses in this section report SimCast-S2S results under this choice.

\begin{figure}[t]
    \centering
    \includegraphics[width=1\linewidth]{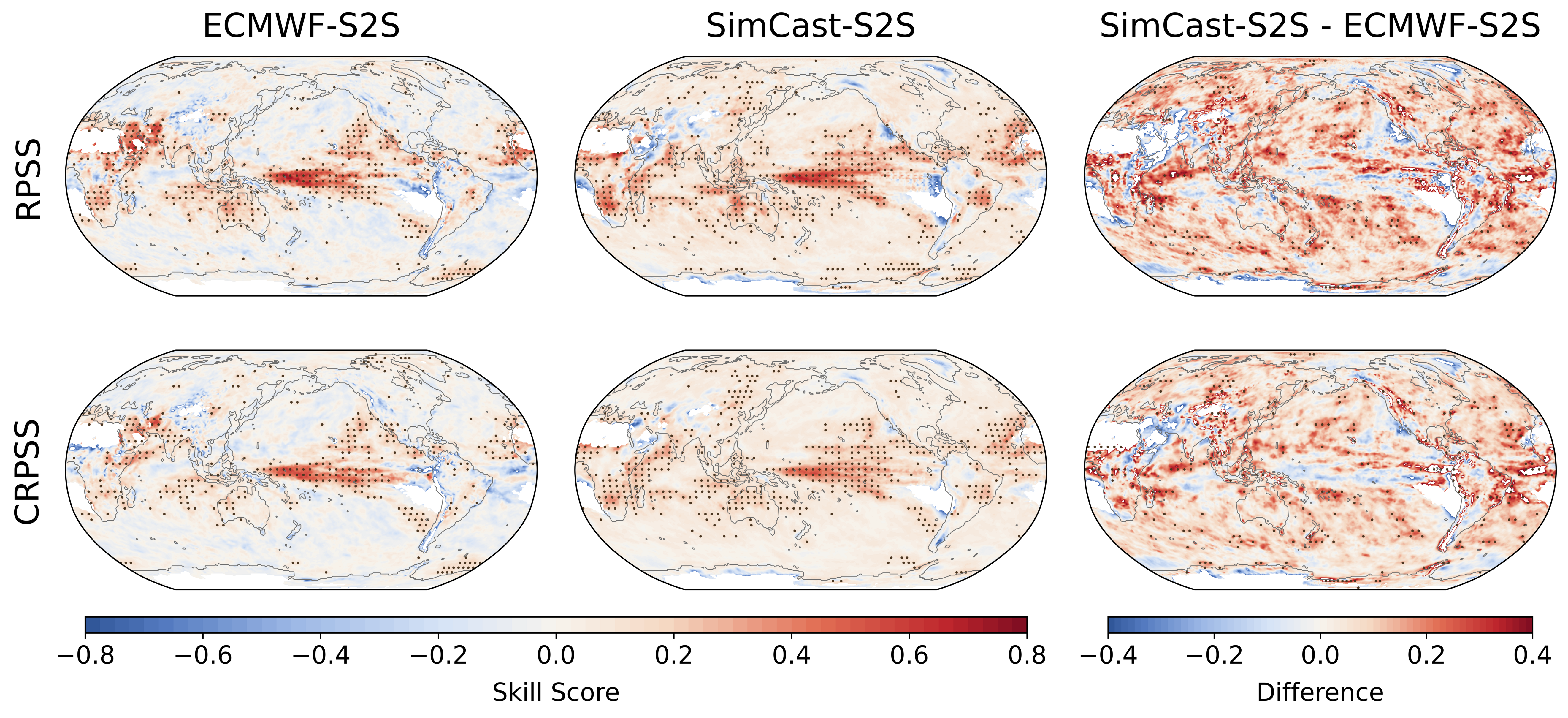}
    \caption{
    Probabilistic forecast skill of ECMWF-S2S and SimCast-S2S for precipitation anomalies.
    \textbf{Top row:} Spatial distribution of tercile-based Ranked Probability Skill Score (RPSS).
    \textbf{Bottom row:} Continuous Ranked Probability Skill Score (CRPSS).
    For each metric, the left and middle columns show the skill scores of ECMWF-S2S and SimCast-S2S, respectively, while the right column shows the difference between SimCast-S2S and ECMWF-S2S.
    Positive values in the left and middle columns indicate skill above the climatological baseline, whereas negative values indicate performance below it.
    Positive values in the right column indicate higher skill for SimCast-S2S relative to ECMWF-S2S.
    Dots indicate regions where the corresponding positive skill score, or positive difference in skill score, is statistically significant at the $2.5\%$ level based on 1,000 bootstrap resamples.
    }
    \label{fig:prob_metrics}
\end{figure}

\begin{figure}[t]
    \includegraphics[width=1.\linewidth]{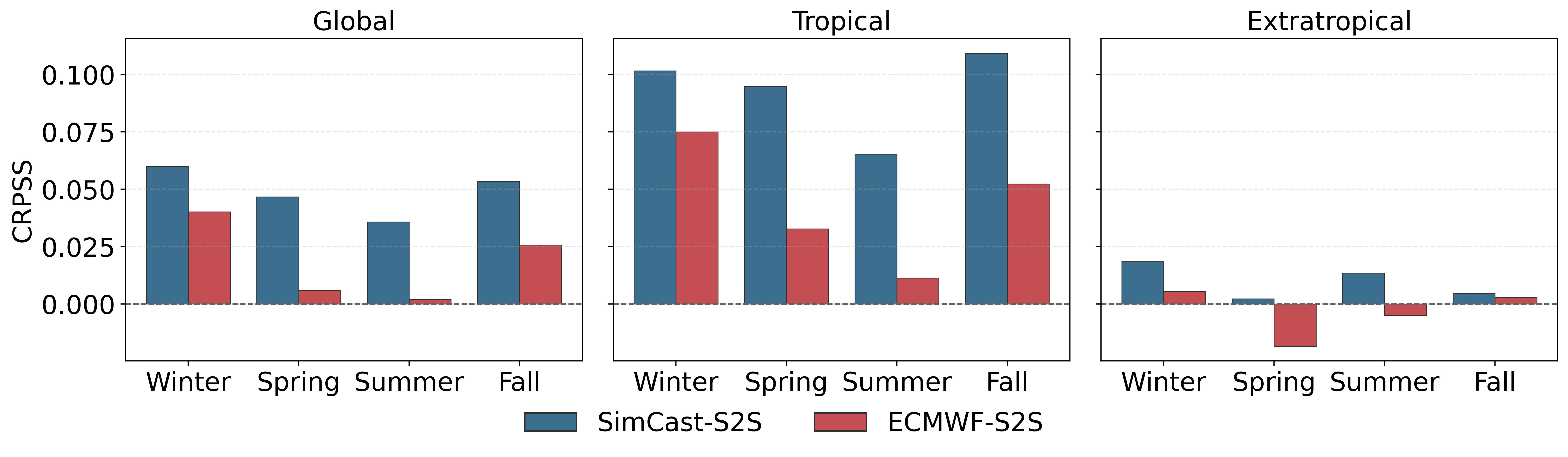}
    \caption{
    Seasonal CRPSS for SimCast-S2S and ECMWF-S2S precipitation forecasts during the 2021--2025 test period over the global, tropical, and extratropical domains. Seasons are defined as winter (December--February), spring (March--May), summer (June--August), and fall (September--November). Positive CRPSS indicates better probabilistic skill than the climatological baseline. Higher values indicate better performance.
    }
    \label{fig:seasonal_crpss}
\end{figure}

\subsection{Probabilistic Skill Evaluation}
\label{subsec:prob}

Deterministic skill evaluation is limiting for subseasonal forecasts because small perturbations in the initial state can lead to substantially different precipitation outcomes; thus, a point forecast can at best approximate the expected value of the conditional distribution and may differ substantially from any individual realization because of the system’s inherent internal variability. Since SimCast-S2S is inherently stochastic, it produces a distribution of possible forecasts rather than a point estimate. We next evaluate how well the forecast distribution produced by SimCast-S2S covers the observed ERA5 target.

Probabilistic forecast skill is strongest in the tropics for both ECMWF-S2S and SimCast-S2S, particularly along the equatorial Pacific, as measured by the ranked probability skill score (RPSS) for tercile forecasts (see Fig.~\ref{fig:prob_metrics}). SimCast-S2S not only exhibits stronger tropical skill pattern but also shows more widespread increased skill relative to ECMWF-S2S across many extratropical regions, especially in the North Pacific, North Asia, North America, the North Atlantic, and the Southern Ocean. This advantage is more clearly visible in the difference map, where positive values indicate higher skill for SimCast-S2S (see red shading). 
The continuous ranked probability skill score (CRPSS) in Fig.~\ref{fig:prob_metrics}, which evaluates the full predictive distribution rather than discrete tercile categories, shows a broadly consistent spatial pattern. SimCast-S2S exhibits higher CRPSS than ECMWF-S2S across most of the globe, indicating that its advantage extends beyond tercile-category prediction to the continuous forecast distribution. It should also be noted that some persistently dry regions, such as the Sahara, the southeastern Pacific off South America, and Antarctica, have precipitation values that are zero or near zero for much of the time, with low temporal variability. In these regions, the verifying precipitation anomalies are small, and both models' forecasts remain close to the reference climatology.

The spatial patterns in Fig.~\ref{fig:prob_metrics} show where probabilistic skill is concentrated. Fig.~\ref{fig:seasonal_crpss} examines which season this skill occurs. Spring and summer appear to be the most difficult seasons to predict subseasonal precipitation variability. ECMWF-S2S shows negative CRPSS over the extratropics in these seasons; however, SimCast-S2S still manages to maintain positive CRPSS. In fact, SimCast-S2S shows higher CRPSS than ECMWF-S2S across all seasons and regions, with the largest differences occurring in the tropics. This agrees with the spatial maps in Fig.~\ref{fig:prob_metrics}, where the strongest probabilistic skill is concentrated along the tropical regions. In the extratropics, both models exhibit weaker skill.

\begin{figure}[t]
    \centering
    \includegraphics[width=1.\linewidth]{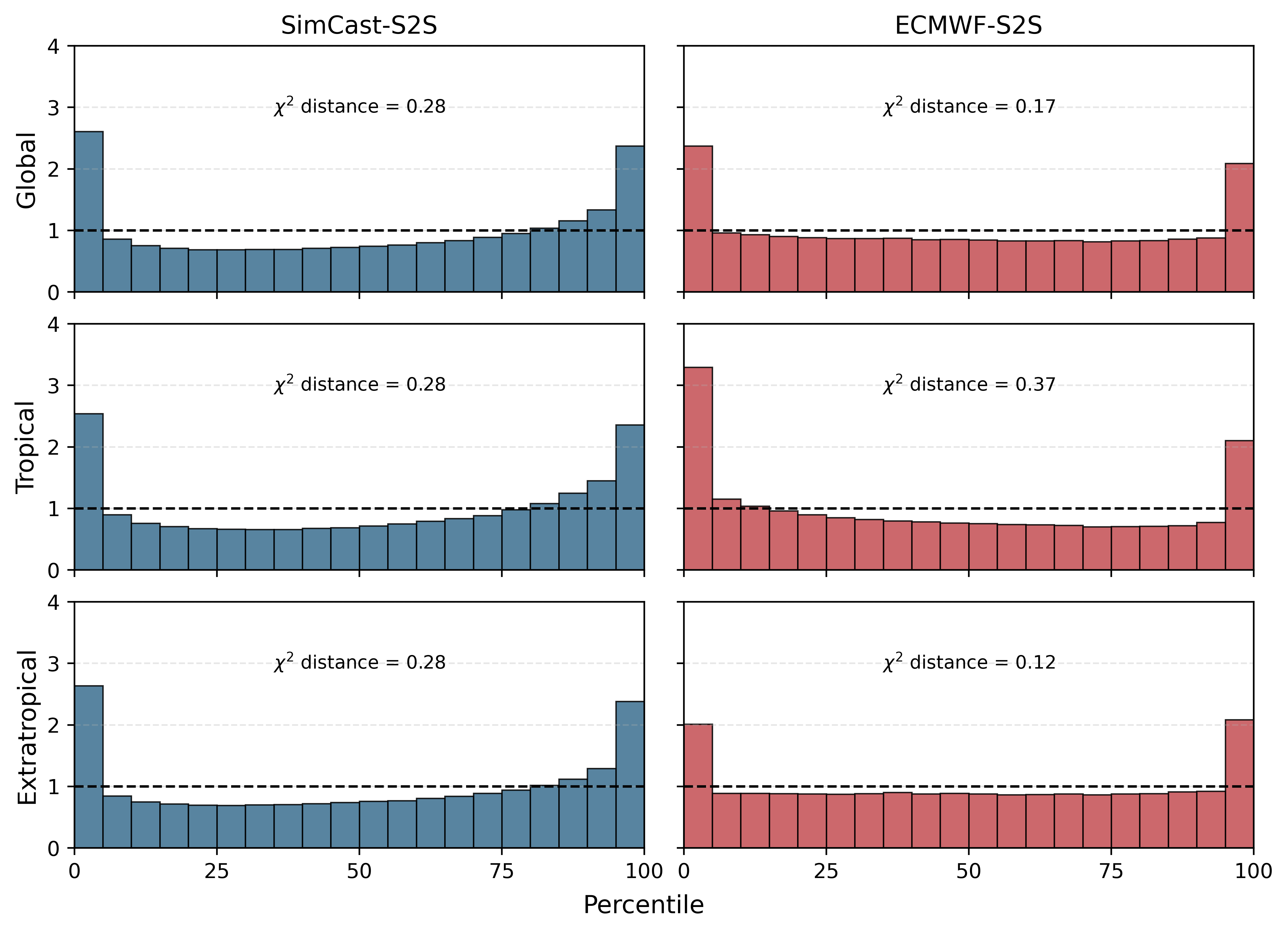}
    \caption{Probability integral transform (PIT) histograms for SimCast-S2S and ECMWF-S2S precipitation forecasts. The red dashed line indicates the uniform distribution expected under perfect calibration. Both models show U-shaped histograms, indicating underdispersive ensemble forecasts. SimCast-S2S shows nearly constant $\chi^2$ distance across regions, whereas ECMWF-S2S shows better extratropical calibration but larger tropical deviation from uniform behavior.}
    \label{fig:pit}
\end{figure}

\subsection{Uncertainty Quantification}
\label{subsec:uncertain}

\begin{figure}[t]
    \centering
    \includegraphics[width=1\linewidth]{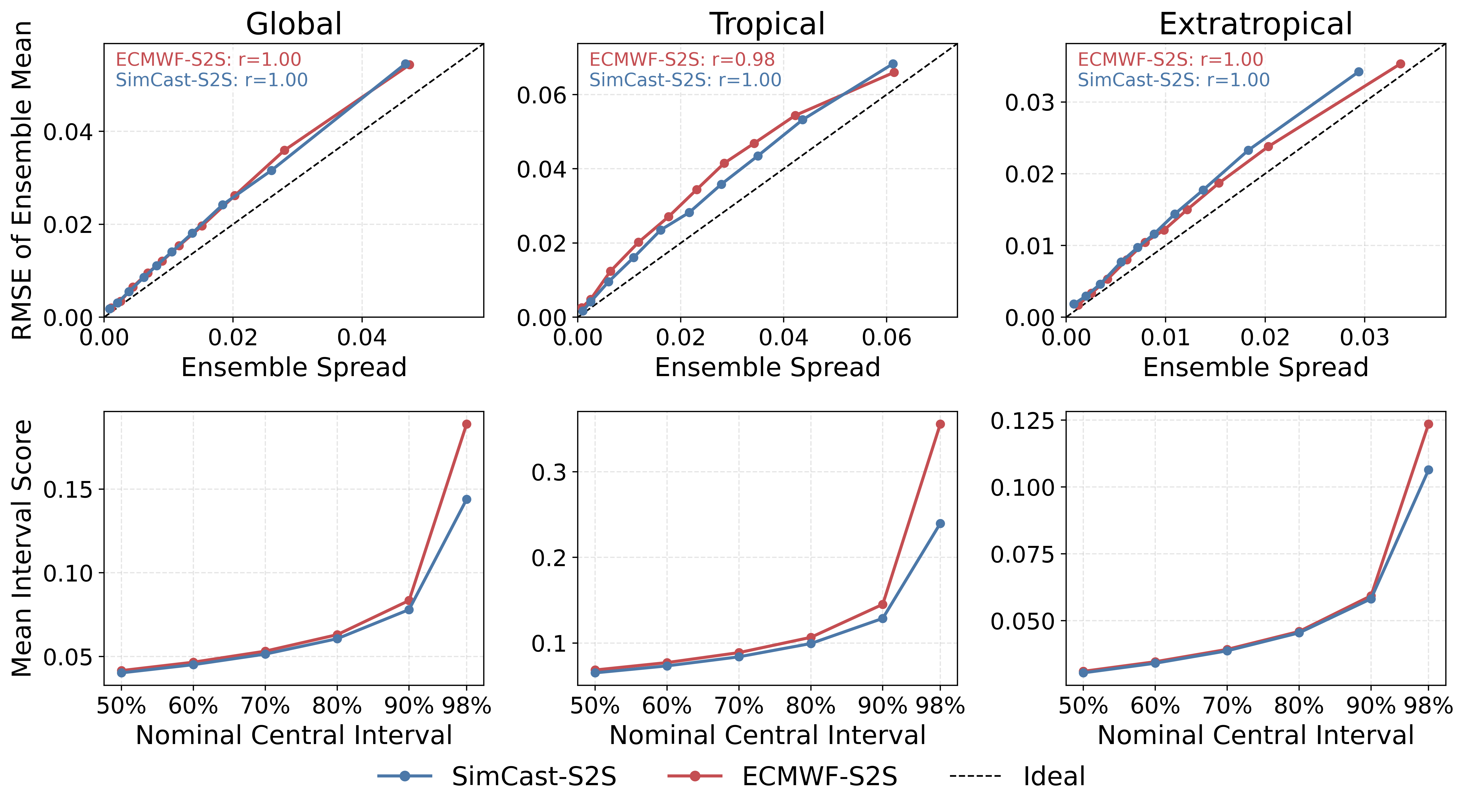}
    \caption{Spread--error and interval-score diagnostics for SimCast-S2S and ECMWF-S2S precipitation forecasts over global, tropical, and extratropical domains. The \textbf{top row} compares mean ensemble spread with the RMSE of the ensemble mean. Both models are above the ideal one-to-one dashed line, indicating forecast errors are larger than ensemble spread. Spread--error correlations are high. This suggests that larger spread is still associated with larger error and the ensemble spread truly reflects useful uncertainty information. The \textbf{bottom row} shows mean interval scores for different central interval coverages, where lower values indicate better probabilistic forecasts. SimCast-S2S consistently exhibits better score, particularly in tropical regions.}
    \label{fig:spread_interval}
\end{figure}

Beyond aggregate probabilistic skill, it is important to examine how well each model represents forecast uncertainty. A useful ensemble forecast should not only produce accurate probabilistic predictions but also assign uncertainty in a way that is statistically consistent with the reanalysis data. In this section, we evaluate uncertainty quantification from two complementary perspectives. First, we examine the percentile histogram to assess the calibration of the ensemble distribution across global, tropical, and extratropical regions. Second, we analyze the relationship between ensemble spread and forecast error, together with interval scores for different nominal coverage levels.

A well-calibrated probabilistic forecast should produce an approximately uniform probability integral transform (PIT) distribution, indicating that the verifying observed/reanalysis estimates are evenly distributed across the forecast distribution. Both SimCast-S2S and ECMWF-S2S exhibit U-shaped PIT histograms, with elevated frequencies near the lowest and highest percentiles across the global, tropical, and extratropical regions (see Fig.~\ref{fig:pit}). This pattern indicates underdispersion, meaning that the forecast distributions are narrow and assign insufficient probability to outcomes in the tails. In order words, both SimCast-S2S and ECMWF-S2S tend to underestimate forecast uncertainty.

The degree of underdispersion varies by region and model. Globally, ECMWF-S2S shows a smaller $\chi^2$ distance from uniform behavior than SimCast-S2S. This global difference is mainly driven by the extratropics, where ECMWF-S2S has a lower $\chi^2$ distance. In the tropics, however, ECMWF-S2S has a higher $\chi^2$ distance than SimCast-S2S. That being said, this comparison should be interpreted carefully. SimCast-S2S is evaluated directly from its raw ensemble output, without additional statistical calibration or post-processing. ECMWF-S2S, in contrast, is accompanied by post-processed forecasts that are commonly used to adjust systematic forecast biases and construct calibrated anomaly or probabilistic products. Therefore, the ECMWF-S2S ensemble evaluated here may not represent an equally raw forecast product in the same sense as in SimCast-S2S. The relatively stable $\chi^2$ distance across regions for SimCast-S2S suggests more spatially uniform ensemble dispersion, whereas the larger tropical--extratropical contrast for ECMWF-S2S indicates stronger regional dependence signaling regional overfitting in the post-processing calibration.

Another key diagnostic of ensemble reliability is the relationship between ensemble spread and forecast error. In a well-scaled ensemble, these two quantities should be close to the one-to-one line, so that larger forecast uncertainty corresponds to larger realized error. Both SimCast-S2S and ECMWF-S2S lie slightly above this line in all three regions of focus, indicating that the realized errors are larger than the ensemble spread (see the top row of Fig.~\ref{fig:spread_interval}). Although high spread--error correlations are observed in both models, the main difference is that SimCast-S2S lies closer to the one-to-one line in most cases over the tropics, and hence provides a better-scaled estimate of forecast uncertainty than ECMWF-S2S in these regions. Over the extratropics, SimCast-S2S slightly underestimates the uncertainty compared to ECMWF-S2S.

The ensemble prediction intervals are further evaluated using the mean interval score (see the bottom row of Fig.~\ref{fig:spread_interval}). The x-axis shows the nominal central interval coverage. For example, a $90\%$ central interval spans the 5th to 95th percentiles of the ensemble distribution, corresponding to the central $90\%$ of the forecast distribution around the median. The y-axis shows the mean interval score. This score penalizes two types of behavior: intervals that are unnecessarily wide and intervals that fail to contain the ground truth. Therefore, a lower interval score indicates both better sharpness and coverage. A formal description of mean interval score can be found in Section \label{subsec:eval}. As expected, the interval score increases as the nominal coverage becomes larger because higher coverage requires wider prediction intervals. Across the global, tropical, and extratropical regions, SimCast-S2S shows slightly lower error than ECMWF-S2S, particularly at the highest coverage levels over the tropical regions. This is consistent with the PIT histogram in Fig. \ref{fig:pit}.

\subsection{Spatial Structure Realism}
\label{subsec:real}

\begin{figure}[t]
    \centering
    \includegraphics[width=1\linewidth]{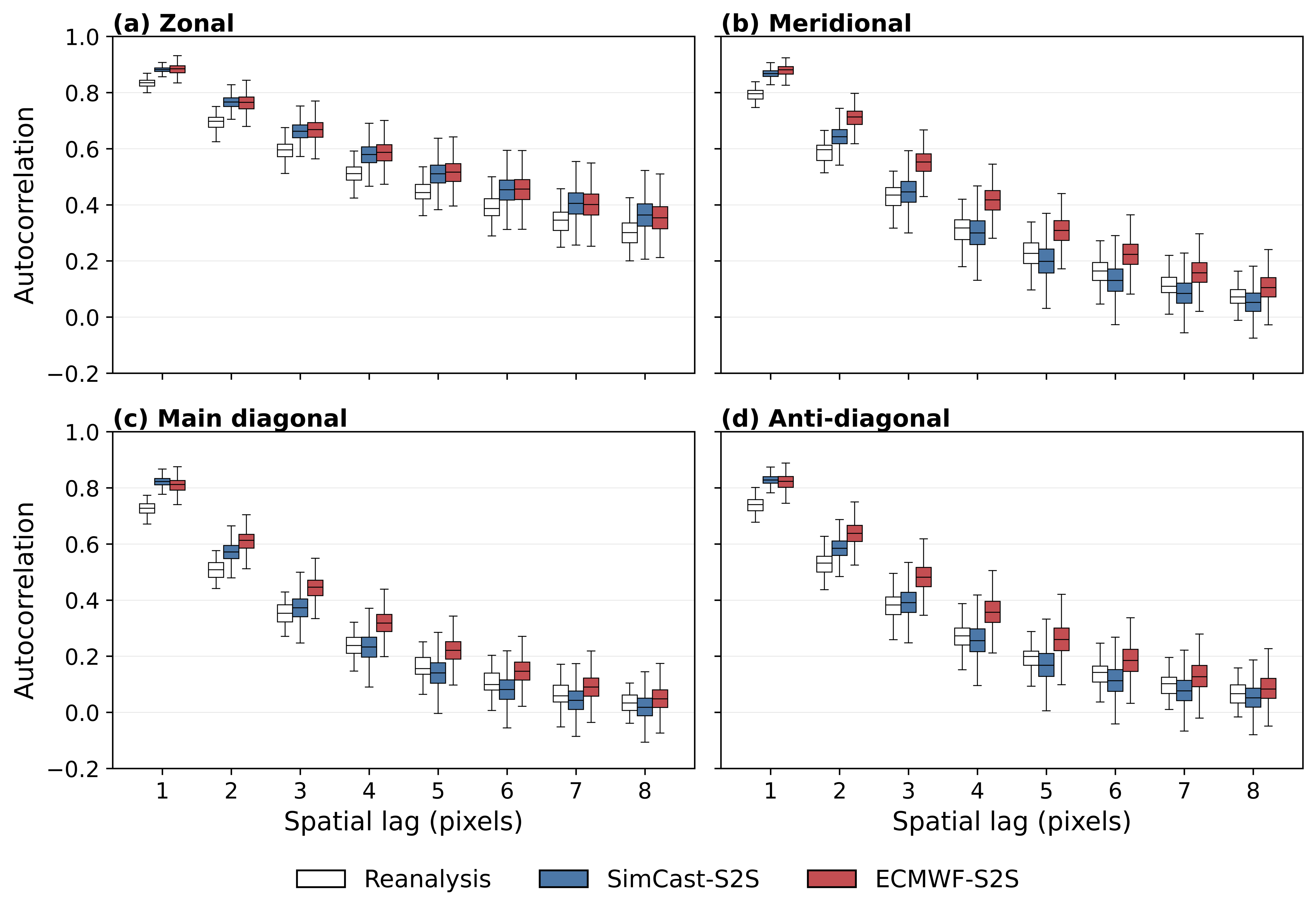}
    \caption{Spatial autocorrelation of ERA5, SimCast-S2S, and ECMWF-S2S precipitation fields along zonal (west--east), meridional (south--north), main-diagonal (northwest--southeast), and anti-diagonal (southwest--northeast) directions. The x-axis shows spatial lag, and the y-axis shows autocorrelation. Smaller differences from the ERA5 distributions indicate better agreement with the ground truth regarding spatial structure of precipitation. SimCast-S2S better follows the true meridional and diagonal autocorrelation, especially at smaller spatial lags where fine-scale precipitation structure remains meaningful and most impactful.}
    \label{fig:realism}
\end{figure}

We next examine whether the forecasted precipitation fields have realistic spatial structure. This is important because a model can achieve good aggregate scores while still producing fields that are too smooth, too noisy, or spatially inconsistent with true precipitation. Spatial autocorrelation is evaluated in four directions: zonal, meridional, main diagonal, and anti-diagonal (see Fig.~\ref{fig:realism}). The meridional direction is especially important because it provides a strong test of whether the models capture realistic transitions across latitude-dependent precipitation regimes.

Across all four directions, the ERA5 fields show decreasing autocorrelation with increasing spatial lag, and both SimCast-S2S and ECMWF-S2S are able to reproduce this basic decay. In the zonal direction, SimCast-S2S and ECMWF-S2S show nearly equal performance. SimCast-S2S and ECMWF-S2S both track the observed decay reasonably well although both models show slightly stronger west-east persistence than the reanalysis data. The meridional direction provides a more demanding test because precipitation regimes vary strongly with latitude, spanning e.g., deep convective precipitation tied to the intertropical convergence zone in the tropics, suppressed precipitation beneath the subsiding branch of the Hadley circulation in the subtropics, frontal and baroclinic precipitation along midlatitude storm tracks, and weaker orographically-driven precipitation in the high-latitudes. In this direction, SimCast-S2S follows the observed decay more closely than ECMWF-S2S across most lags whereas ECMWF-S2S unfavorably persists more spatial structures across latitude bands.

The main diagonal and anti-diagonal directions follow the northwest--southeast and southwest--northeast directions, respectively. In both directions, SimCast-S2S is closer to the true autocorrelation than ECMWF-S2S at smaller spatial lags, where local precipitation structure is still meaningfully present. As the lag increases, the autocorrelation quickly collapses to zero. This is the distance range where precipitation at one location has little statistical connection to precipitation farther away. Consequently, the advantage of SimCast-S2S diminishes (although still present), not because the models become equally realistic, but because there is little remaining spatial dependence for either model to reproduce.

\subsection{Computational Efficiency and Scalability}
\label{subsec:eff}

A practical advantage of SimCast-S2S is that probabilistic forecasts can be generated rapidly once the model is deployed.  Conventional subseasonal systems such as ECMWF-S2S generate probabilistic forecasts by repeatedly integrating a full numerical model forward in time for many ensemble members. SimCast-S2S instead produces ensemble members through stochastic sampling during the denoising process of the diffusion model (see Subsection \ref{subsec:simcasts2s}), so each member can be generated independently and parallelized efficiently across GPUs.

\begin{figure}
    \centering
    \includegraphics[width=1\linewidth]{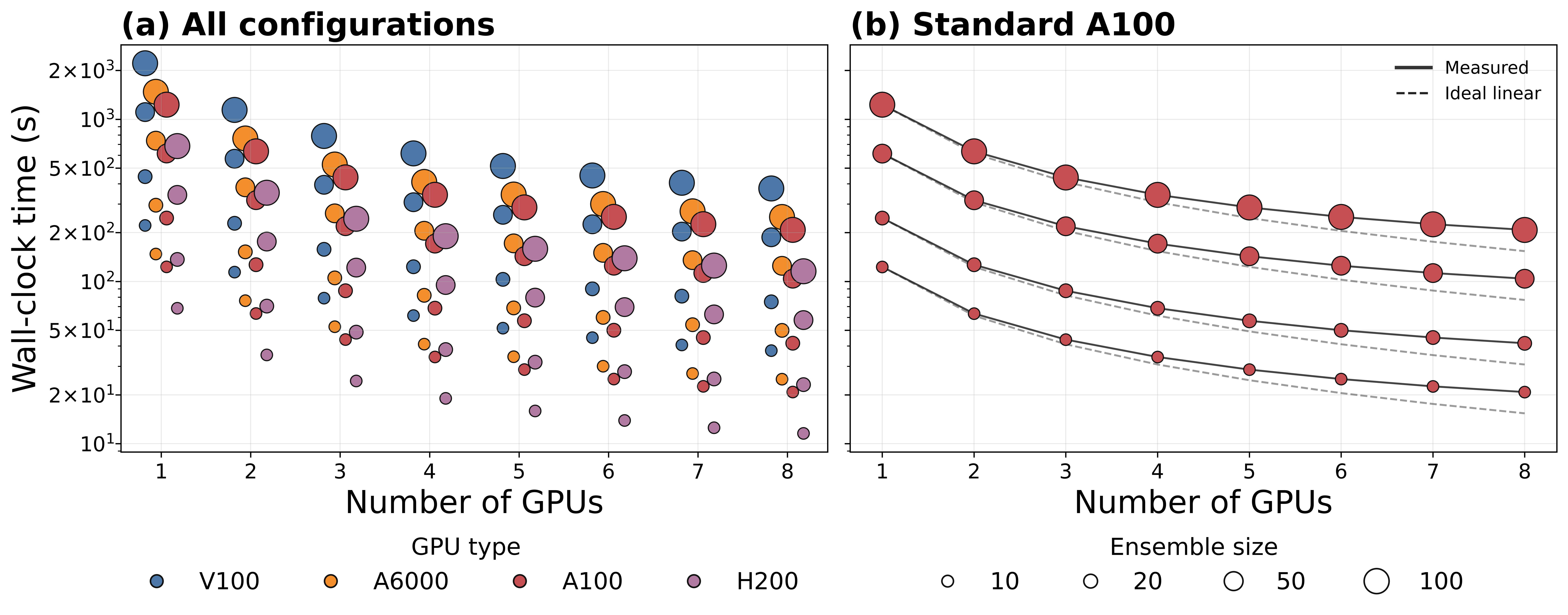}
    \caption{Computational efficiency and scalability of SimCast-S2S ensemble generation. \textbf{Panel~(a)} shows wall-clock inference time as a function of number of GPUs, GPU types and ensemble sizes. \textbf{Panel~(b)} compares measured A100 runtimes with ideal linear scaling; closer agreement with the dashed lines indicates better multi-GPU scalability.}
    \label{fig:eff}
\end{figure}

SimCast-S2S can generate large precipitation ensembles efficiently at $192 \times 288$ resolution. On a single A100 GPU, it takes approximately 12.3 seconds per ensemble member, or 123 seconds for 10 members, 246 seconds for 20 members, 615 seconds for 50 members, and 1230 seconds, or 20.5 minutes, for 100 members (see Fig.~\ref{fig:eff}). With three A100 GPUs, the same 100-member ensemble requires 438.21 seconds, or about 7.3 minutes. With four A100 GPUs, it decreases further to 342.19 seconds, or about 5.7 minutes. Thus, a 100-member ensemble can be generated in less than ten minutes using a small number of industry-grade A100 GPUs.

It should be noted that the scaling behavior is close to linear because ensemble generation is highly parallelizable. Moving from one to two A100 GPUs reduces the 100-member runtime from 1230.0 seconds to 634.19 seconds, corresponding to a parallel efficiency of $\frac{1230}{2 \times 634.19} = 0.9697$. With three GPUs, the efficiency is 0.9356, and with four GPUs it is 0.8986. The efficiency decreases gradually as more GPUs are added, reaching 0.7392 at eight GPUs (see panel~(b) of Fig.~\ref{fig:eff}). This departure from ideal linear scaling is small and expected~\cite{li2020pytorch,amdahl1967validity}. It reflects fixed overheads from model loading, data movement, GPU synchronization, and inter-device communication. These overheads are more visible when the ensemble size is small, but they are increasingly amortized as the number of ensemble members grows~\cite{gustafson1988reevaluating}. SimCast-S2S also benefits directly from newer accelerator hardware. For a 100-member ensemble on one GPU, the wall-clock time is 2213 seconds on V100, 1477 seconds on A6000, 1231 seconds on A100, and 683 seconds on H200 (see panel~(a) of Fig.~\ref{fig:eff}). With eight H200 GPUs, the 100-member runtime decreases to only 115.55 seconds, or less than two minutes. This is a strong indication that SimCast-S2S can exploit both per-GPU improvement and multi-GPU parallelism.

To our knowledge, ECMWF does not publicly disclose the wall-clock integration time. Products corresponding to the 00:00 UTC forecast base time are scheduled to become available at 20:00 UTC~\cite{ecmwf_subseasonal_set_vi}. However, this 20-hour interval likely encompasses an end-to-end operational workflow, including initialization, numerical integration, post-processing, and dissemination, not just model runtime. An earlier ECMWF benchmark nevertheless provides some indication of the computational expense. Its previous 51-member, 15-day ensemble required an average wall-clock time of 82 minutes on 1,530 compute nodes~\cite{bauer2020scalability}. Although this historical benchmark is not directly comparable with the runtime of SimCast-S2S, it illustrates the large-scale computational resources required by conventional dynamical ensemble forecasting. In contrast, generating a 100-member probabilistic ensemble with SimCast-S2S is just a matter of minutes on a single GPU node. This efficiency makes large-ensemble S2S prediction widely accessible to typical research groups and also allows uncertainty estimates and extreme-event probabilities to be produced with much lower latency.

\section{Discussion}
\label{sec:discussion}
We tackle three key bottlenecks that have limited data-driven models for S2S forecasting. First, generative models are notoriously data-hungry; we address this through transfer learning via LoRA, pretraining on large ensembles of climate simulations before fine-tuning on limited reanalysis data. Second, generating large probabilistic ensembles at high resolution is computationally prohibitive; we address this by operating in a compact latent space rather than physical space, which makes large-ensemble generation efficient. Third, S2S forecasting fundamentally requires representing forecast uncertainty, not just a single deterministic outcome; we address this by using a generative (diffusion-based) framework that directly models the forecast distribution rather than a point estimate. We show that each of these design choices works: our model outperforms all deterministic AI baselines, and using only a subset of input variables and without any post-processing or bias correction, it challenges a leading operational forecasting system.

A central result of this study is that the higher probabilistic skill of SimCast-S2S also translates into stronger deterministic performance. SimCast-S2S exhibits higher probabilistic skill across much of the globe as shown in Figs.~\ref{fig:prob_metrics} and~\ref{fig:seasonal_crpss}, consistent with the lower deterministic errors reported in Table~\ref{tab:mae_precip}. This result suggests that the generated forecast members are not merely adding stochastic variability around a weak deterministic forecast; rather, they contain a coherent predictive signal that survives ensemble averaging and is closer to the true conditional distribution.

Our uncertainty diagnostics in Subsection~\ref{subsec:uncertain} indicate that both SimCast-S2S and ECMWF-S2S remain underdispersive, i.e., their ensemble spread is narrow relative to realized true variability. However, the degree and regional structure of this underdispersion differ. ECMWF-S2S shows stronger uncertainty behavior in the extratropics, while SimCast-S2S is more competitive in the tropics and it overall lies closer to the spread--RMSE $1{:}1$ reference. This comparison is demanding for SimCast-S2S because ECMWF-S2S uses reforecasts to estimate model-specific systematic errors; bias correction then reduces persistent offsets, and probabilistic calibration adjusts ensemble probabilities toward observed event frequencies. In contrast, SimCast-S2S is evaluated directly from its raw ensemble output without comparable post-processing or calibration. Therefore, its competitive tropical performance and improved spread--error scaling suggest that the generative ensemble already contains meaningful uncertainty information before any statistical correction is applied.

Importantly, we show that SimCast-S2S better captures the spatial decay of precipitation dependence, especially in the meridional and diagonal directions. This is notable because precipitation regimes vary strongly with latitude, from tropical convection to subtropical suppression, midlatitude storm tracks and high-latitude stratiform precipitation. It is well known that deep convolutional forecast models trained with point-wise loss functions favor overly smooth fields and suppress fine-scale precipitation structure~\citep{mathieu2016deep, ravuri2021skillful, harris2022generative}. In contrast, SimCast-S2S learns the spatial organization of precipitation across dynamically distinct regimes. This is one of the strongest arguments for using a diffusion-based generative model in this setting. The model is trained to sample from a learned distribution of atmospheric states, and therefore can preserve spatial dependence in ways that cannot be done by point-wise loss functions alone.

SimCast-S2S also changes the practical cost structure of subseasonal ensemble prediction. Conventional dynamical systems generate ensemble forecasts by repeatedly integrating a full numerical model forward in time. SimCast-S2S directly models the one-to-one mapping, from the current atmospheric state to the target weeks. This is done through neural-network stochastic sampling, where individual members can be generated independently and parallelized efficiently across accelerators. As shown in Subsection~\ref{subsec:eff}, a 100-member ensemble can be generated in minutes on a small GPU node. The same ensemble size would require thousands of CPU nodes and several hours of wall-clock time in a conventional dynamical forecasting system~\cite{bauer2020scalability}. This efficiency opens opportunities for research settings where operational-scale computing is unavailable, and for applications that require many ensemble members, such as uncertainty quantification, tail-risk assessment, scenario generation, and impact modeling~\citep{gneiting2014probabilistic, kurth2023fourcastnet, price2025probabilistic}.

A limitation is that the realism of SimCast-S2S is currently statistical rather than explicitly physical. The model can learn realistic distributions, spatial structures, and predictor--target relationships, but it is not guaranteed to satisfy conservation laws or dynamical balances. This limitation is closely tied to the same design choice that makes the model efficient: latent-space sampling. Working in a low-dimensional latent representation largely reduces computational cost, but physical laws such as mass conservation, moisture conservation, and momentum balance only make sense in physical space. There is no obvious way to impose these laws directly on abstract latent variables. As a result, SimCast-S2S may generate fields that are statistically plausible but not strictly dynamically consistent. A natural next step is therefore to combine latent-space generation with (occasional) physical-space correction. One possible direction is a hybrid sampling scheme in which the model generates most of the trajectory in latent space, periodically decodes the state into physical variables, evaluates physical residuals, applies corrections, and then maps the corrected state back into latent space. For precipitation forecasting, this could involve constraints derived from the moisture budget, physical bounds on humidity and precipitation, or consistency between circulation and moisture convergence. Such a scheme is feasible because neural networks are inherently differentiable, and physical residuals can be computed directly from their computation graph~\citep{baydin2018automatic, raissi2019physics, karniadakis2021physics,nguyen2023parc,nguyen2024parcv2}. Incorporating physical constraints into the generative model may also help improve representation of extremes, which remain a key weakness of both SimCast-S2S and ECMWF-S2S.

Another important direction is interpretability. SimCast-S2S uses many atmospheric predictors, but the present analysis does not fully explain which fields, regions, or dynamical patterns the model relies on when generating forecasts. Explainable AI methods could help determine whether the model uses physically meaningful sources of predictability, such as tropical moisture anomalies, upper-level circulation, midlatitude wave activity, atmospheric rivers, or teleconnection patterns~\citep{mamalakis2022xai, mamalakis2022fidelity, higgins2023deep, higgins2024subseasonal, singh2023role, singh2023innovative, goyal2024understanding, corner2026shortwave}. Methods such as input perturbation, integrated gradients, latent-space sensitivity analysis, and counterfactual sampling could be adapted to our generative setting~\citep{ribeiro2016why, sundararajan2017axiomatic, wachter2017counterfactual}. This would be especially valuable because generative models can be right for the wrong reasons, and such failures are difficult to diagnose from forecast scores alone. Therefore, connecting SimCast-S2S skill to interpretable physical mechanisms becomes crucial for making the model more scientifically useful and for identifying regimes in which it is likely to fail.

Our results demonstrate the potential of latent, diffusion-based, generative modeling combined with simulation-to-reanalysis transfer learning as an efficient and scalable framework for a new generation of probabilistic S2S precipitation forecasts. Further advances in physical consistency and interpretability could strengthen their reliability and scientific utility.

\section{Methods}
\label{sec:method}

\subsection{Data}
\label{subsec:data}

SimCast-S2S is designed to maximize computational efficiency so that it can be trained on large ensembles of climate-model simulations and then transferred to observation-constrained reanalysis data. We first pretrain the model on 28 ensembles from the Community Earth System Model version 2 (CESM2), a fully coupled Earth system model that represents interactions among the atmosphere, ocean, land surface, sea ice, and other climate components \cite{danabasoglu2020cesm2,rodgers2021cesm2le}. After pretraining on CESM2 simulations, SimCast-S2S is fine-tuned and evaluated on the fifth-generation European Centre for Medium-Range Weather Forecasts atmospheric reanalysis, ERA5 \cite{hersbach2020era5}. ERA5 combines a global numerical weather prediction model with a large collection of real-world observations through data assimilation, including information from satellites, radiosondes, aircraft, and surface stations. Therefore, ERA5 is not a strictly observational dataset, but rather an observation-constrained reconstruction of the atmosphere. As a result, it provides one of the closest available approximations to the observed atmospheric state and can be used to transfer the model from idealized climate-model simulations to realistic atmospheric conditions.

To make transfer learning between CESM2 and ERA5 physically meaningful, the input variables must be defined consistently across the two datasets. We therefore select variables that are available, or have close physical counterparts, in both CESM2 and ERA5. The selected variables are also chosen to describe the atmospheric column across three representative pressure levels: 850 hPa for the lower troposphere, 500 hPa for the middle troposphere, and 200 hPa for the upper troposphere. In addition, single-level variables are included to describe surface and column-integrated conditions that are directly linked to precipitation, including surface temperature, sea-level pressure, outgoing longwave radiation, total precipitable water, and precipitation itself.

The list of all CESM2 variables used for pretraining and their ERA5 counterparts used for fine-tuning are provided in Table~\ref{tab:variables}. At each pressure level, we include wind, geopotential, temperature, and humidity. These variables jointly describe the dynamical and thermodynamical state of the atmosphere: winds capture horizontal and vertical transport, geopotential represents large-scale circulation structure, temperature describes thermal stratification, and humidity provides the moisture supply needed for precipitation. At the surface or column-integrated level, the selected variables provide additional constraints on boundary-layer conditions, radiative state, pressure patterns, and atmospheric moisture content.

\begin{table}[t]
\centering
\caption{CESM2 input variables used for SimCast-S2S pretraining and their corresponding ERA5 variables used for fine-tuning.}
\begin{tabular}{llll}
\toprule
\textbf{Pressure level} & Variable type & CESM2 variable & ERA5 variable \\
\midrule

\multirow{5}{*}{200 hPa}
& Zonal wind          & U200 & u\_200 \\
& Meridional wind     & V200 & v\_200 \\
& Temperature         & T200 & t\_200 \\
& Specific humidity   & Q200 & q\_200 \\
& Geopotential height & Z200 & z\_200 \\

\midrule

\multirow{6}{*}{500 hPa}
& Zonal wind          & U500 & u\_500 \\
& Meridional wind     & V500 & v\_500 \\
& Vertical velocity   & OMEGA500 & w\_500 \\
& Temperature         & T500 & t\_500 \\
& Specific humidity   & Q500 & q\_500 \\
& Geopotential height & Z500 & z\_500 \\

\midrule

\multirow{5}{*}{850 hPa}
& Zonal wind          & U850 & u\_850 \\
& Meridional wind     & V850 & v\_850 \\
& Temperature         & T850 & t\_850 \\
& Specific humidity   & Q850 & q\_850 \\
& Geopotential height & Z850 & z\_850 \\

\midrule

\multirow{5}{*}{\textbf{Single level}}
& Surface temperature & TS & t2m \\
& Sea-level pressure  & PSL & msl \\
& Precipitation       & PRECT & tp \\
& Top longwave flux   & FLUT & avg\_tnlwrf \\
& Total precipitable water & TMQ & tcwv \\

\bottomrule
\end{tabular}
\label{tab:variables}
\end{table}

Some variables are physically related but not identical across the two datasets, so simple conversions are required. For example, ERA5 reports average top net longwave radiation flux (\texttt{avg\_tnlwrf}), following the ECMWF sign convention in which downward flux is positive. This has the opposite sign convention from the CESM2 upwelling longwave flux at the top of the model (\texttt{FLUT}), so the ERA5 field is multiplied by $-1$. Similarly, CESM2 reports geopotential height (\texttt{Z}), whereas ERA5 reports geopotential (\texttt{z}). To make the variables equivalent, ERA5 geopotential is divided by gravitational acceleration, $g \approx 9.80665~\mathrm{m~s^{-2}}$, to obtain geopotential height. Precipitation also requires unit adjustment. ERA5 reports accumulated precipitation over an hourly interval (\texttt{tp}), whereas CESM2 reports total precipitation as a per-second rate (\texttt{PRECT}). Therefore, ERA5 precipitation is divided by $3,600$. These conversions are made to ensure that SimCast-S2S is pretrained and fine-tuned on variables with comparable units.

All input and target variables were transformed into standardized anomalies before model training. From a meteorological standpoint, this step separates subseasonal variability from the much stronger background structure of the climate system, including the seasonal cycle, spatially varying climatology, and slow long-term trends. Raw atmospheric fields are dominated by predictable differences between seasons and regions; however, subseasonal forecasting is primarily concerned with departures from those expected states. Removing the local trend and day-of-year climatology therefore encourages the model to learn physically meaningful anomaly relationships rather than simply reproducing the mean annual cycle or the background climate state. The same transformation was applied to both input variables and the precipitation target to preserve a consistent meteorological reference frame. In this form, the model learns how anomalous large-scale atmospheric conditions, such as circulation, temperature, and moisture anomalies, are associated with anomalous precipitation. Specifically, for each variable, a grid-point-wise linear trend was first estimated from the daily fields and removed to reduce slow nonstationary changes unrelated to subseasonal variability. A smoothed day-of-year climatology was then obtained. After that, each field was standardized by subtracting the corresponding climatological mean and dividing by the climatological standard deviation. For validation and testing, the detrending coefficients and climatological statistics estimated from the training period were reused to avoid information leakage.

\subsection{Physical and Latent Space}
\label{subsec:vae}

SimCast-S2S represents the atmospheric state in two coupled spaces: the gridded physical space and a learned probabilistic latent space. In physical space, each sample is a collection of standardized anomaly fields defined on a latitude--longitude grid, with separate variable groups describing circulation, mass, thermodynamic, moisture, and precipitation processes. Although this representation preserves direct meteorological interpretability, it is high-dimensional and highly redundant because large-scale atmospheric fields exhibit strong spatial covariance and coherent dynamical structures. Directly modeling the conditional distribution of future precipitation in this space would therefore require learning stochastic relationships across a large number of mutually dependent grid points.

\begin{figure}
\centering
    \includegraphics[width=1\linewidth]{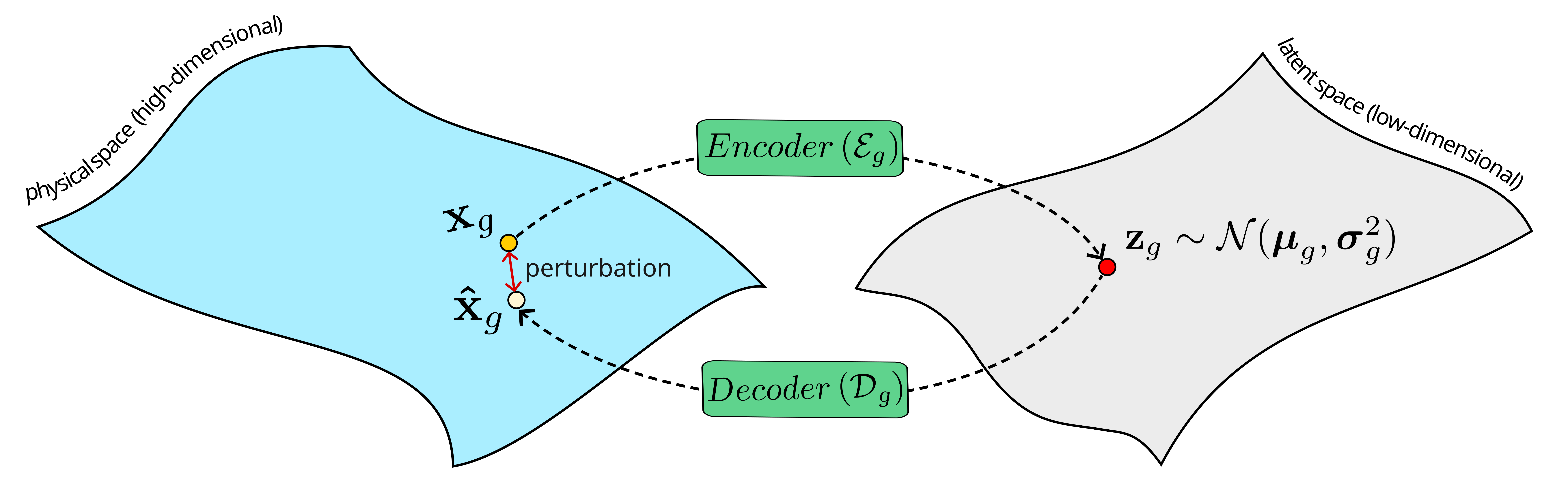}
    \caption{Schematic illustration of the VAE mapping between the high-dimensional physical space and the low-dimensional latent space for variable group $g$. The encoder $\mathcal{E}_g$ maps a anomaly field $\mathbf{x}_g$ to a Gaussian latent distribution $\mathbf{z}_g \sim \mathcal{N}(\boldsymbol{\mu}_g, \boldsymbol{\sigma}_g^2)$. Sampling from this distribution introduces stochasticity in latent space, and the decoder $\mathcal{D}_g$ maps the sampled latent representation back to a reconstructed physical field $\hat{\mathbf{x}}_g$. The latent space has a much lower dimension than the physical space, and stochastic sampling introduces a small perturbation between $\hat{\mathbf{x}}_g$ and $\mathbf{x}_g$.}
\label{fig:vae}
\end{figure}

To reduce dimensionality, SimCast-S2S uses variational autoencoders (VAEs) \cite{kingma2014autoencoding} to map each physical variable group from its gridded physical space to a compact probabilistic latent space. For variable group $g$, let $\mathbf{x}_g \in \mathcal{X}_g \subset \mathbb{R}^{C_g \times H \times W}$ denote a physical-space field group, where $C_g$ is the number of variables in the group and $H \times W$ is the latitude--longitude grid. The corresponding latent representation is defined in a lower-dimensional space $\mathbf{z}_g \in \mathcal{Z}_g \subset \mathbb{R}^{d_g}$, where $d_g \ll C_gHW$. Under a parameterization $\phi$, the encoder therefore learns a probabilistic mapping between the high-dimensional physical space $\mathcal{X}_g$ and the compact latent space $\mathcal{Z}_g$, i.e. $\mathcal{E}_{\phi_g}: \mathcal{X}_g \to \mathcal{Z}_g$. 

Although the encoder $\mathcal{E}_{\phi_g}$ models a probabilistic latent representation, it is still a deterministic neural network. Given an input field group $\mathbf{x}_g$, the encoder first predicts the mean and standard deviation of the approximate posterior:

\begin{equation}
\left(\boldsymbol{\mu}_g, \boldsymbol{\sigma}_g^2 \right) = \mathcal{E}_{\phi_g} \left[ \mathbf{x}_g \right].
\end{equation}

These parameters define a Gaussian approximate posterior $q_{\phi}\left( \mathbf{z}_g \mid \mathbf{x}_g \right) = \mathcal{N} \left(\boldsymbol{\mu}_g,\, \operatorname{diag}\left(\boldsymbol{\sigma}_g^2\right)\right).$ The stochasticity is introduced by sampling $\mathbf{z}_g$ from this approximate posterior. To allow gradients to pass through, the sampling is done using a continuous reparameterization trick:

\begin{equation}
    \mathbf{z}_{g} = \boldsymbol{\mu}_{g} + \boldsymbol{\sigma}_{g} \odot \boldsymbol{\epsilon},
\end{equation}

where $\odot$ is the element-wise multiplication, and $\boldsymbol{\epsilon} \sim \mathcal{N}(\mathbf{0}, \mathbf{I})$. 

The decoder $\mathcal{D}_{\theta_g}: \mathcal{Z}_g \to \mathcal{X}_g$, parameterized by $\theta$, maps the sampled latent vector back to physical space:

\begin{equation}
\hat{\mathbf{x}}_g = \mathcal{D}_{\theta_g}\left(\mathbf{z}_g\right).
\end{equation}

where $\hat{\mathbf{x}}_g \in \mathcal{X}_g$ is the reconstructed standardized anomaly field. 

In probabilistic form, the decoder parameterizes the conditional reconstruction distribution $p_{\theta}\left(\mathbf{x}_g \mid \mathbf{z}_g\right)$, which describes the distribution of physical-space fields that can be reconstructed from a given latent state $\mathbf{z}_g$.

There is always some discrepancy between $\hat{\mathbf{x}}_g$ and $\mathbf{x}_g$ for two reasons. First, the latent vector $\mathbf{z}_g \sim q_{\phi}\left( \mathbf{z}_g \mid \mathbf{x}_g \right)$ is a random variable, different samples from the same approximate posterior can lead to slightly different decoded fields. Second, some information is inevitably lost when a high-dimensional physical field is compressed into a much lower-dimensional latent space. As a result, the same anomaly field $\mathbf{x}_g$, when passed through the encoding and decoding processes multiple times, can produce multiple possible reconstructions $\hat{\mathbf{x}}_g$. This introduces a natural source of perturbation in the reconstructed physical fields, which supports the stochastic nature of SimCast-S2S. Fig.~\ref{fig:vae} illustrates this encoding--decoding process.

The learning objective of a VAE is to maximize the marginal likelihood of the observed data $p_{\phi,\theta}(\mathbf{x})$. Although this marginal likelihood is intractable, it can be showed that this objective can be obtained equivalently by minimizing its negative evidence lower bound (ELBO), which leads to the following loss function:

\begin{equation}
\label{eq:loss_vae}
    \mathcal{L}_{\text{VAE}}(\boldsymbol{\theta}, \boldsymbol{\phi}, \mathbf{x}) := 
    - \mathbb{E}_{q_{\boldsymbol{\phi}}(\mathbf{z} \mid \mathbf{x})} [\log p_{\boldsymbol{\theta}}(\mathbf{x} \mid \mathbf{z})] 
    + D_{\mathrm{KL}}[ q_{\boldsymbol{\phi}}(\mathbf{z} \mid \mathbf{x}) \,\|\, p(\mathbf{z})].
\end{equation}

The first term in Eq. \ref{eq:loss_vae} is the negative log-likelihood of the observed dataset, which encourages the model to reconstruct the data accurately. The second term is the KL divergence between two distributions, which acts as a regularizer that encourages the approximate posterior $q_{\boldsymbol{\phi}}(\mathbf{z} \mid \mathbf{x})$ to remain close to the prior $p(\mathbf{z})$. 

In practice, it is common to further assume that the reconstruction error follows a Gaussian likelihood, i.e., $\mathbf{x} - \hat{\mathbf{x}} \sim \mathcal{N}(\mathbf{0}, \sigma^2 \mathbf{I})$, where and $\sigma \in \mathbb{R}$ is a fixed constant. This is equivalent to modeling the likelihood as $p_\theta(\mathbf{x} \mid \mathbf{z}) = \mathcal{N}(\mathbf{x}; \, \hat{\mathbf{x}}, \sigma^2 \mathbf{I})$. Under this assumption, the negative log-likelihood simplifies to a scaled MSE loss, and minimizing the negative log-likelihood is equivalent to minimizing the MSE between the reconstruction and the input, up to a constant scaling factor.

\begin{equation}
\label{eq:neg_loglikelihood}
    -\log p_\theta(\mathbf{x} \mid \mathbf{z}) = \frac{1}{2\sigma^2} \|\mathbf{x} - \hat{\mathbf{x}}\|^2 + \text{const}.
\end{equation}

Also, with the prior $p(\mathbf{z})$ assumed to be a standard Gaussian, the KL divergence in Eq. \ref{eq:loss_vae} becomes a special case and can be written in a compact form:

\begin{equation}
\label{eq:kl_divergence}
    D_{\mathrm{KL}} \left[ q_{\boldsymbol{\phi}}(\mathbf{z} \mid \mathbf{x}) \,\|\, p_{\boldsymbol{\theta}}(\mathbf{z}) \right] = \frac{1}{2} \sum_{j=1}^{d_g} \left( \sigma_j^2 + \mu_j^2 - 1 - \log \sigma_j^2 \right),
\end{equation}

where $\mu_i$ and $\sigma_i$ are the components of $\boldsymbol{\mu}_g$ and $\boldsymbol{\sigma}_g$, respectively.

\begin{figure}
\centering
    \includegraphics[width=1\linewidth]{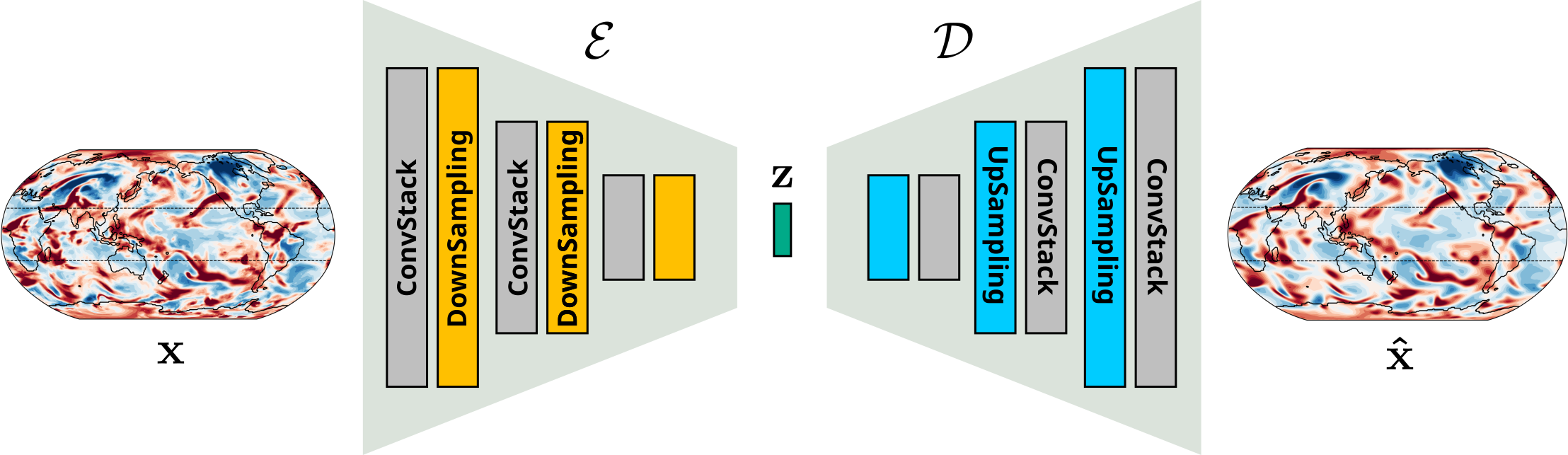}
    \caption{
    VAE architecture used to map physical-space anomaly fields into a compact latent representation and reconstruct them back to physical space. The encoder $\mathcal{E}$ progressively compresses the input field $\mathbf{x}$ through ConvStack and DownSampling blocks to obtain the latent vector $\mathbf{z}$. The decoder $\mathcal{D}$ mirrors this process using UpSampling and ConvStack blocks to reconstruct the field $\hat{\mathbf{x}}$.
    }
\label{fig:vae_architecture}
\end{figure}

Combining Eq. \ref{eq:loss_vae}, \ref{eq:kl_divergence}, \ref{eq:neg_loglikelihood}, the training procedure of VAE is summarized in Algo. \ref{algo:vae_training}. The encoder parameters $\boldsymbol{\phi}$ and decoder parameters $\boldsymbol{\theta}$ are optimized jointly using the Adam optimizer~\citep{kingma2015adam}. Fig.~\ref{fig:vae_architecture} describe the structure of $\mathcal{E}_{\phi_g}$ and $\mathcal{D}_{\theta_g}$, which together form the complete VAE architecture. Both the encoder and decoder use a standard convolutional architecture~\citep{lecun1998gradient, krizhevsky2012imagenet}. The encoder is built from stacked convolutional layers, each followed by a downsampling block to progressively reduce the spatial resolution by a factor for four. The decoder mirrors this structure but uses upsampling blocks. Each upsampling block contains a transposed convolutional layer to expand the spatial resolution by a factor for four. 

The number of ConvStack--DownScaling blocks directly controls the latent dimension. Let $C_g$ denote the number of physical variables in group $g$, let $E_g$ denote the number of latent channels per variable, and let $N_g$ denote the number of ConvStack--DownScaling blocks, the latent dimension is $d_g = \frac{C_g E_g H W}{4^{N_g}}$, where $HW$ is the original physical-space grid. The corresponding compression ratio is $\frac{4^{N_g}}{E_g}$. For example, for a group of $C_g = 7$ variables, setting $N_g = 3$ and $E_g = 28$ gives a $16\times$ compression. For a group of $C_g = 1$ variable, setting $N_g = 3$ and $E_g = 16$ gives a $4\times$ compression.

\begin{algorithm}[t]
\caption{VAE Training Procedure for variable group $g$}
\begin{algorithmic}[0]
\Require Dataset $\mathbb{D}$, encoder $\mathcal{E}_{\phi_g}$, decoder $\mathcal{D}_{\theta_g}$, loss balancing term $\lambda \in \mathbb{R}^{+}$
\Repeat
    \State Sample $\mathbf{x} \in \mathbb{D}$
    \State Encoder: compute $\left(\boldsymbol{\mu}_g, \boldsymbol{\sigma}_g^2 \right) = \mathcal{E}_{\phi_g} \left[ \mathbf{x}_g \right]$
    \State Sample $\boldsymbol{\epsilon} \sim \mathcal{N}(\mathbf{0}, \mathbf{I})$
    \State Compute $\mathbf{z}_{g} = \boldsymbol{\mu}_{g} + \boldsymbol{\sigma}_{g} \odot \boldsymbol{\epsilon}$
    \State Decoder: compute $\hat{\mathbf{x}}_{g} = \mathcal{D}_{\theta_{g}}\left(\mathbf{z}_{g}\right)$
    \State Take a gradient step on:
    \Statex \hspace{20pt}
    $\nabla_{\theta,\phi}
    \left[
    \lambda \|\mathbf{x} - \hat{\mathbf{x}}\|^2 + \frac{1}{2} \sum_{j=1}^{d_g} \left( \sigma_j^2 + \mu_j^2 - 1 - \log \sigma_j^2 \right)
    \right]$
\Until{converged}
\end{algorithmic}
\label{algo:vae_training}
\end{algorithm}

\begin{table}
\centering
\caption{
Variable groups used in SimCast-S2S and their corresponding CESM2 and ERA5 variables. The 21 input variables are divided into five physically motivated groups: wind, mass, thermal, hydro, and precipitation. Each group has physical-space dimension $C_gHW$, latent-space dimension $d_g$, and compression ratio $C_gHW/d_g$. Here, $H \times W = 192 \times 288$.
}
\label{tab:variable_groups}
\small
\setlength{\tabcolsep}{4pt}
\renewcommand{\arraystretch}{1.15}
\begin{tabular}{
>{\raggedright\arraybackslash}p{0.09\textwidth}
>{\raggedright\arraybackslash}p{0.27\textwidth}
>{\raggedright\arraybackslash}p{0.27\textwidth}
ccc}
\toprule
\textbf{Group}
& \textbf{CESM2 variables}
& \textbf{ERA5 variables}
& $\boldsymbol{C_gHW}$
& $\boldsymbol{d_g}$
& \textbf{Compression} \\
\midrule

wind
& U200, V200, U500, V500, OMEGA500, U850, V850
& u200, v200, u500, v500, w500, u850, v850
& $387{,}072$
& $24{,}192$
& $16$ times \\
\midrule

mass
& Z200, Z500, Z850, PSL
& z200, z500, z850, msl
& $221{,}184$
& $13{,}824$
& $16$ times \\
\midrule

thermal
& TS, T200, T500, T850, FLUT
& skt, t200, t500, t850, avgtnlwrf
& $276{,}480$
& $17{,}280$
& $16$ times \\
\midrule

hydro
& TMQ, Q200, Q500, Q850
& tcwv, q200, q500, q850
& $221{,}184$
& $13{,}824$
& $16$ times \\
\midrule

precip
& PRECT
& tp
& $55{,}296$
& $13{,}824$
& $4$ times \\

\bottomrule
\end{tabular}
\end{table}

In this study, 21 variables listed in Table~\ref{tab:variables} are divided into five physically motivated groups, namely \texttt{wind}, \texttt{mass}, \texttt{thermal}, \texttt{hydro}, and \texttt{precip}. Table~\ref{tab:variable_groups} details the variables included in each group for both CESM2 and ERA5. Each group has a physical-space dimension of $C_gHW$ where $HW = 192 \times 288 = 55{,}296$. Choosing the compression level involves a trade-off between efficiency and reconstruction accuracy: a smaller latent dimension gives a more compact representation, while a larger latent dimension reduces reconstruction error. Empirically, we find reconstruction error decreases approximately exponentially as the latent dimension increases. Based on our analysis, we choose a $16\times$ compression for the wind, mass, thermal, and hydro groups, and a $4\times$ compression for precipitation. The lower compression ratio for precipitation is necessary because its distribution is highly intermittent, strongly skewed, and dominated by localized extremes. More aggressive compression would therefore oversmooth the latent distribution and degrade reconstruction fidelity.

\subsection{SimCast-S2S model}
\label{subsec:simcasts2s}

SimCast-S2S is a latent diffusion model~\citep{sohl2015deep, song2019generative, ho2020denoising, song2021scorebased, rombach2022high, yang2023diffusion}. The central idea is to model the stochastic evolution in the low-dimensional latent space. During the forward diffusion process, Gaussian noise is gradually added to the target latent vector until it is fully corrupted into a standard Gaussian. A neural network is then trained to reverse this process: starting from a noisy target latent state, it progressively removes noise by conditioning on the latent representations of the past atmospheric states. The objective is to recover the original target latent representation.

Consider a variance schedule $0 < \beta_1, \beta_2, \cdots, \beta_K < 1$. For a precipitation latent $\mathbf{z}_0 \sim p(\mathbf{z})$, 
a discrete Markov chain $\{ \mathbf{z}_0, \mathbf{z}_1, \cdots, \mathbf{z}_K \}$ is defined by:

\begin{equation}
\label{eq:ddpm_fwd}
    \mathbf{z}_k := \sqrt{1 - \beta_t} \, \mathbf{z}_{k-1} + \sqrt{\beta_k} \, \mathbf{\boldsymbol{\varepsilon}}_k, \quad \mathbf{\boldsymbol{\varepsilon}}_k \sim \mathcal{N}(\mathbf{0}, \mathbf{I}), \quad k: 1 \to K,
\end{equation}

or equivalently:

\begin{equation}
\label{eq:ddpm_fwd_derived}
q(\mathbf{z}_k \mid \mathbf{z}_{k-1}) = \mathcal{N}\left( \mathbf{z}_k; \sqrt{1 - \beta_k} \, \mathbf{z}_{k-1}, \beta_k \mathbf{I} \right).
\end{equation}

Eq. \ref{eq:ddpm_fwd} defines a first-order Markov chain where the signal is gradually degraded by $\sqrt{1 - \beta_k}$, and Gaussian noise with variance $\beta_k$ is injected at each step. As $k \to K$, $\mathbf{z}_K \sim \mathcal{N}(\mathbf{0}, \mathbf{I})$.

Because each transition is Gaussian and linear, we can compute $q(\mathbf{z}_k \mid \mathbf{z}_0)$ in closed form. Defining $\bar{\alpha}_k := \prod_{j=1}^k (1 - \beta_j)$ where $\bar{\alpha}_0 = 1$, which quantifies how much signal remains from $\mathbf{z}_0$ at step $k$, Eq.~\ref{eq:ddpm_fwd} is now simplified as:

\begin{equation}
\label{eq:ddpm_fwd_simple}
    \mathbf{z}_k = \sqrt{\bar{\alpha}_k} \mathbf{z}_0 + \sqrt{1 - \bar{\alpha}_k} \, \boldsymbol{\epsilon},
\end{equation}

in which $\boldsymbol{\epsilon} \sim \mathcal{N}(\mathbf{0}, \mathbf{I})$ is a fresh standard Gaussian variable. This implies that $q(\mathbf{z}_k \mid \mathbf{z}_0)$ is Gaussian with a closed form:

\begin{equation}
\label{eq:ddpm_reparam}
    q(\mathbf{z}_k \mid \mathbf{z}_0) = \mathcal{N}(\mathbf{z}_k; \sqrt{\bar{\alpha}_k} \, \mathbf{z}_0, (1 - \bar{\alpha}_k) \mathbf{I}),
\end{equation}

Eq. \ref{eq:ddpm_reparam} is important because it enables direct sampling of $\mathbf{x}_k$ from the clean data point $\mathbf{x}_0$ without iteratively going through all the diffusion steps.

We can train a neural network $f_{\theta}(\mathbf{Z}_\mathrm{cond}, t, \mathbf{z}_k, k)$ to approximate $\boldsymbol{\epsilon}$, where $\mathbf{Z}_\mathrm{cond}$ is the concatenated latent representation of the initial physical state and $t$ is the day of year as detailed in section~\ref{subsec:vae}. The training objective reduces to a simple $\ell_2$ loss function:

\begin{equation}
    \label{eq:ddpm_loss}
    \mathcal{L}_\text{SimCast-S2S}(\theta) = \mathbb{E}_{\mathbf{z}_0, t} \left[ \left\| \boldsymbol{\epsilon} - f_{\theta}(\boldsymbol{Z}_\mathrm{cond}, t, \mathbf{z}_k, k) \right\|_2^2 \right],
\end{equation}

Because the forward process $q(\mathbf{z}_{k} \mid \mathbf{z}_{k-1})$ is Markov and Gaussian, it can be shown that the reverse process $p(\mathbf{z}_{k-1} \mid \mathbf{z}_{k})$ is also Markov and Gaussian:

\begin{equation}
    p_\theta(\mathbf{z}_{k-1} \mid \mathbf{z}_k) = \mathcal{N}\left( \mathbf{z}_{k-1}; \boldsymbol{\mu}_\theta(\boldsymbol{Z}_\mathrm{cond}, t, \mathbf{z}_k, k), \, \Sigma_{k} \mathbf{I} \right).
\end{equation}

The mean $\boldsymbol{\mu}_\theta(\boldsymbol{Z}_\mathrm{cond}, t, \mathbf{z}_k, k)$ is not a separate neural network by itself, but is analytically computed from the noise prediction $\boldsymbol{\hat{\epsilon}}_{k-1} = f_{\theta}(\boldsymbol{Z}_\mathrm{cond}, t, \mathbf{z}_k, k)$. In practice, the variance $\Sigma_{k} = \mathrm{Var}[\mathbf{x}_{k-1} \mid \mathbf{x}_{k}]$ is either fixed or learned. In this study, we make a common choice of setting $\Sigma_{k} = \tilde{\beta}_k = \frac{1 - \bar{\alpha}_{k-1}}{1 - \bar{\alpha}_k} \beta_k$. This choice ensures the reverse variance matches the true posterior variance $\tilde{\beta}_{k} = \mathrm{Var}[\mathbf{z}_{k-1} \mid \mathbf{z}_k, \mathbf{z}_0]$.

\begin{figure}
    \centering
    \includegraphics[width=1\linewidth]{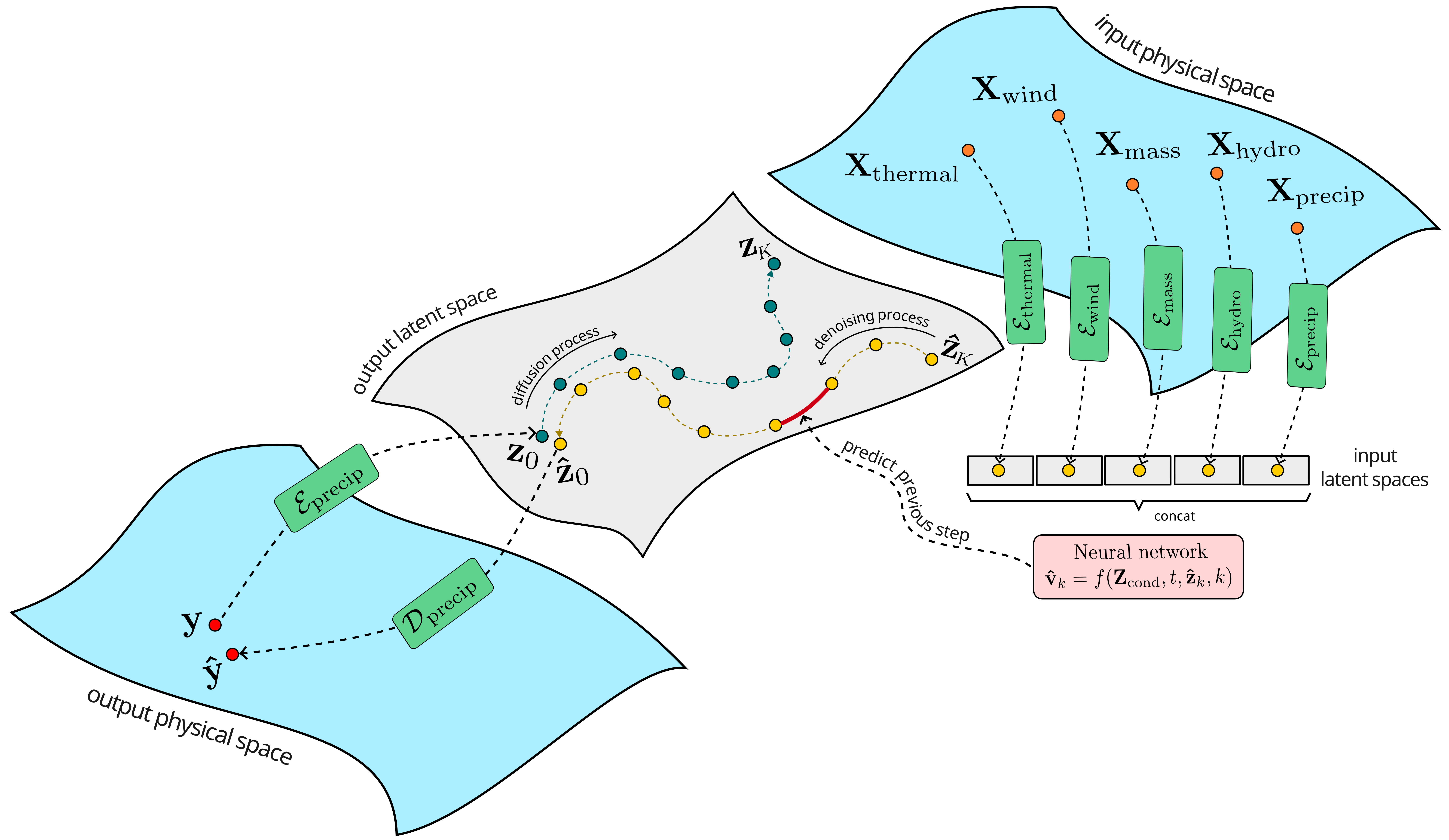}
    \caption{
    Schematic illustration of the SimCast-S2S latent diffusion framework. Historical atmospheric fields are first separated into five physical groups: wind, thermal, mass, hydro, and precipitation, as described in Table~\ref{tab:variable_groups}. Each group is encoded by its pretrained VAE encoder into a low-dimensional latent space. The resulting conditioning latents are concatenated to form $\mathbf{Z}_{\mathrm{cond}}$. Starting from a random Gaussian latent $\hat{\mathbf{z}}_K \sim \mathcal{N}(\mathbf{0}, \mathbf{I})$, the neural network iteratively denoises the latent state, conditioned on $\mathbf{Z}_{\mathrm{cond}}$, the denoising step $k$, and seasonal information $t$, until it produces the target precipitation latent $\hat{\mathbf{z}}_0$. The pretrained precipitation decoder $\mathcal{D}_{\mathrm{precip}}$ then maps $\hat{\mathbf{z}}_0$ back to physical space to obtain the subseasonal precipitation-anomaly forecast field $\hat{\mathbf{y}}$.
    }
\label{fig:diffusion}
\end{figure}

Recall that the generative process $p_\theta(\mathbf{z}_{k-1} \mid \mathbf{z}_k)$ is meant to approximate the true reverse process of the forward Markov chain. The ideal reverse transition is the posterior distribution $q(\mathbf{z}_{k-1} \mid \mathbf{z}_k)$, but this is intractable. However, we can exactly compute the posterior $q(\mathbf{z}_{k-1} \mid \mathbf{z}_k, \mathbf{z}_0)$, which is the distribution over $\mathbf{z}_{k-1}$ conditioned on both the noisy input $\mathbf{z}_k$ and the original clean sample $\mathbf{z}_0$. Here, we use $q(\mathbf{z}_{k-1} \mid \mathbf{z}_k, \mathbf{z}_0)$ as a tractable surrogate for the intractable marginal posterior $q(\mathbf{z}_{k-1} \mid \mathbf{z}_k)$. Because affine transformations and marginalizations of Gaussians preserve Gaussianity, the triplet $(\mathbf{z}_0, \mathbf{z}_{k-1}, \mathbf{z}_k)$ jointly forms a multivariate Gaussian distribution, and conditioning on any subset—including both $\mathbf{z}_0$ and $\mathbf{z}_k$—also yields another Gaussian. This allows us to analytically compute the posterior $q(\mathbf{z}_{k-1} \mid \mathbf{z}_k, \mathbf{z}_0)$ in closed form using standard results from Gaussian conditioning:

\begin{equation}
\label{eq:ddpm_bwd_surrogate}
    q(\mathbf{z}_{k-1} \mid \mathbf{z}_k, \mathbf{z}_0) = \mathcal{N}\left( \mathbf{z}_{k-1}; \tilde{\boldsymbol{\mu}}(\mathbf{z}_k, \mathbf{z}_0), \Sigma_k \mathbf{I} \right),
\end{equation}

where

\begin{equation}
    \tilde{\boldsymbol{\mu}}(\mathbf{z}_k, \mathbf{z}_0) = \mathbb{E}[\mathbf{z}_{k-1} \mid \mathbf{z}_k, \mathbf{z}_0] = \frac{\sqrt{\bar{\alpha}_{k-1}} \beta_k}{1 - \bar{\alpha}_k} \, \mathbf{z}_0 + \frac{\sqrt{\alpha_k}(1 - \bar{\alpha}_{k-1})}{1 - \bar{\alpha}_k} \, \mathbf{z}_k,
\end{equation}

During the generative process, we do not have access to the true clean sample $\mathbf{z}_0$. To approximate the reverse transition distribution $q(\mathbf{z}_{k-1} \mid \mathbf{z}_k, \mathbf{z}_0)$, we leverage the reparameterized form of the forward process, which expresses $\mathbf{z}_k$ as a linear combination of $\mathbf{z}_0$ and Gaussian noise (see Eq.~\ref{eq:ddpm_fwd_simple}). Using the predicted noise $\boldsymbol{\hat{\epsilon}} = f_{\theta}(\mathbf{Z}_\mathrm{cond}, t, \mathbf{z}_k, k)$, we can estimate $\mathbf{z}_0$ as:

\begin{equation}
\label{eq:ddpm_predicted_x0}
    \hat{\mathbf{z}}_0 := \frac{1}{\sqrt{\bar{\alpha}_k}} \left[ \mathbf{z}_k - \sqrt{1 - \bar{\alpha}_k} \, \boldsymbol{\hat{\epsilon}} \right].
\end{equation}

This estimator effectively inverts the forward process at timestep $k$ under the assumption that $\boldsymbol{\hat{\epsilon}}$ accurately predicts the noise component added to $\mathbf{z}_0$ to obtain $\mathbf{z}_k$. We then substitute this estimate $\hat{\mathbf{z}}_0$ into the analytic expression for the true posterior mean $\tilde{\boldsymbol{\mu}}(\mathbf{z}_k, \mathbf{z}_0)$. With algebraic simplification, this expression becomes a compact and efficient form that is commonly used in practice:

\begin{equation}
    \tilde{\boldsymbol{\mu}}(\mathbf{z}_k, \hat{\mathbf{z}}_0) = \frac{1}{\sqrt{1 - \beta_k}} \left( \mathbf{z}_k - \frac{\beta_k}{\sqrt{1 - \bar{\alpha}_k}} \, \boldsymbol{\hat{\epsilon}} \right).
\end{equation}

With the mean in Eq. \ref{eq:ddpm_bwd_surrogate} now fully defined, the full stochastic reverse process at timestep $k$ is obtained by sampling from a Gaussian centered at this predicted mean and with variance given by $\Sigma_{k} = \tilde{\beta}_k = \frac{1 - \bar{\alpha}_{k-1}}{1 - \bar{\alpha}_k} \beta_k$. This yields the update rule used during sampling:

\begin{equation}
\label{eq:ddpm_update_rule}
    \mathbf{z}_{k-1} = \frac{1}{\sqrt{1 - \beta_k}} \left( \mathbf{z}_{k} - \frac{\beta_k}{\sqrt{1 - \bar{\alpha}_k}} \, \boldsymbol{\hat{\epsilon}}\right) + \sqrt{\Sigma_k} \, \boldsymbol{\epsilon}, \quad \text{where } \boldsymbol{\epsilon} \sim \mathcal{N}(\mathbf{0}, \mathbf{I}),
\end{equation}

which defines one step of the generative chain, gradually transforming pure noise into a clean latent.

Eq.~\ref{eq:ddpm_update_rule} is known as the Denoising Diffusion Probabilistic Models (DDPM)~\cite{ho2020denoising}. In this study, we also employ a method to further generalize this update rule by decomposing the stochasticity in Eq. \ref{eq:ddpm_fwd_simple} into a component along the data-aligned direction $\boldsymbol{\epsilon}_{\theta}$, and a fresh isotropic noise $\boldsymbol{\zeta} \sim \mathcal{N}(\mathbf{0}, \mathbf{I})$:

\begin{equation}
\label{eq:ddim_family_general}
    \mathbf{z}_{k-1} = \sqrt{\bar{\alpha}_{k-1}}\,\hat{\mathbf{z}}_0 + c_k\, \boldsymbol{\hat{\epsilon}} + \sigma_k\, \boldsymbol{\zeta}, 
\end{equation}

with the scalar coefficients $c_k \ge 0$ and $\sigma_k \ge 0$ to be determined. To ensure variance matching, the variance of $\mathbf{z}_{k-1}$ conditioned on $\mathbf{z}_0$ under Eq. \ref{eq:ddim_family_general} must satisfy the constraint:

\begin{equation}
    \mathrm{Var}[\mathbf{x}_{k-1}\mid \mathbf{x}_0] \;=\; c_k^2\,\mathbf{I} \;+\; \sigma_k^2\,\mathbf{I}.
\end{equation}

or equivalently,

\begin{equation}
\label{eq:ddim_coef_constraint}
c_k^2 \;+\; \sigma_k^2 = 1 - \bar{\alpha}_{k-1}
\end{equation}

Among all pairs $(c_k, \sigma_k)$ that satisfy the constraint in Eq. \ref{eq:ddim_coef_constraint}, we introduce a parameter $\eta \in [0, 1]$ to control the split between $c_k$ and $\sigma_k$. This is done by defining

\begin{equation}
\label{eq:ddim_eta_def}
    \sigma_k^2 := \eta^2 \, \tilde{\beta}_k,
\end{equation}

The remaining variance is allocated to $c_k$:

\begin{equation}
\label{eq:ddim_ck_def}
    c_t^2 = 1 - \bar{\alpha}_{t-1} - \sigma_t^2.
\end{equation}

There are two important special cases. When $\eta = 1$, we have $\sigma_k^2 = \tilde{\beta}_k$. It can be shown that this recovers the DDPM sampling rule described in Eq.~\ref{eq:ddpm_update_rule}. When $\eta = 0$, we have $\sigma_k = 0$ and $c_k = \sqrt{1-\bar{\alpha}_{k-1}}$. In this case, the reverse process becomes deterministic and corresponds to Denoising Diffusion Implicit Model (DDIM) sampling~\citep{song2021denoising}. Eq.~\ref{eq:ddim_family_general} then reduces to

\begin{equation}
    \mathbf{z}_{k-1} = \sqrt{\bar{\alpha}_{k-1}} \, \hat{\mathbf{z}}_0 + \sqrt{1-\bar{\alpha}_{k-1}} \, \boldsymbol{\hat{\epsilon}} .
\end{equation}

When $0 < \eta < 1$, we obtain an interpolation between the deterministic path ($\eta=0$) and the fully stochastic path ($\eta=1$):

\begin{equation}
\label{eq:ddim_interpolation}
    \mathbf{z}_{k-1}
    = \sqrt{\bar{\alpha}_{k-1}}\, \hat{\mathbf{z}}_0
    + \sqrt{\,1-\bar{\alpha}_{k-1} - \sigma_k^2\,}\;
      \underbrace{\frac{\mathbf{z}_k - \sqrt{\bar{\alpha}_k}\,\hat{\mathbf{z}}_0}{\sqrt{\,1-\bar{\alpha}_k\,}}}_{\boldsymbol{\hat{\epsilon}}}
    + \sigma_k\, \boldsymbol{\zeta},
    \quad \boldsymbol{\zeta}\sim\mathcal{N}(\mathbf{0},\mathbf{I}),
\end{equation}

in which, $\sigma^2_{t}$ takes the general form:
$\sigma_t^2 = \eta^2 \,\tilde{\beta}_t = \eta^2 \cdot \frac{1-\bar{\alpha}_{t-1}}{\,1-\bar{\alpha}_t\,} \left(1 - \frac{\bar{\alpha}_t}{\bar{\alpha}_{t-1}}\right)$.

The interpolated form from Eq. \ref{eq:ddim_interpolation} has its mean still depending on $\hat{\mathbf{z}}_0$ and $\boldsymbol{\hat{\epsilon}}$, but the variance injected is scaled down by $\eta^2$. Thus, $\eta$ interpolates between two anchors: when $\eta=0$, the process is purely deterministic, when $\eta=1$, we recover the standard stochastic sampling; and for $0 < \eta < 1$, the process offers a trade-off between diversity and stability.

So far, this formulation assumes that the neural network predicts the noise term $\boldsymbol{\epsilon}$ directly. However, this is not the only possible parameterization of the denoising objective. Instead of predicting $\boldsymbol{\epsilon}$, SimCast-S2S is trained to predict a velocity variable $\mathbf{v}$, which represents a rotated coordinate in the two-dimensional plane spanned by the clean latent $\mathbf{z}_0$ and the noise $\boldsymbol{\epsilon}$. This $\mathbf{v}$-prediction formulation provides an alternative way to describe the same diffusion trajectory. However, as will be shown later in this subsection, $\mathbf{v}$-prediction yields a more stable training scheme than $\boldsymbol{\epsilon}$-prediction. 

Recall that the forward process is defined as $\mathbf{z}_k = \sqrt{\bar{\alpha}_k}\,\mathbf{z}_0 + \sqrt{1-\bar{\alpha}_k}\,\boldsymbol{\epsilon}$ (Eq. \ref{eq:ddpm_fwd_simple}), where $\boldsymbol{\epsilon} \sim \mathcal{N}(0, I)$ is pure Gaussian and $\mathbf{z}_0$ denotes the clean data sample. Because $\mathbf{z}_0$ and $\boldsymbol{\epsilon}$ are independent and orthogonal in expectation, i.e. $\mathbb{E}[\mathbf{z}_0^\top \boldsymbol{\epsilon}] = 0$, their covariance vanishes: 

\begin{equation}
\mathrm{Cov}(\mathbf{z}_0,\boldsymbol{\epsilon}) = \mathbb E\!\big[(\mathbf{z}_0-\mathbb E\mathbf{z}_0)^\top(\boldsymbol{\epsilon}-\mathbb E\boldsymbol{\epsilon})\big] = \mathbb E(\mathbf{z}_0-\mathbb E\mathbf{z}_0)^\top\mathbb E(\boldsymbol{\epsilon} - \mathbb{E}\boldsymbol{\epsilon}) = \mathbf{0},
\end{equation}

\begin{figure*}[t]
\centering
\begin{tikzpicture}[scale=2., >=Stealth]
    \def\om{35} 
    \coordinate (O) at (0,0);
    \coordinate (X1) at (1,0);
    \coordinate (Y1) at (0,1);
    \coordinate (P) at ({cos(\om)},{sin(\om)});
    \coordinate (Px) at ({cos(\om)},0);
    \coordinate (Py) at (0,{sin(\om)});
    \draw[-] (-1.1,0) -- (1.15,0);
    \draw[-] (0,-1.1) -- (0,1.15);
    \draw[->] (1.15,0) -- +(0.12,0);
    \draw[->] (0,1.15) -- +(0,0.12);
    \draw[line width=0.8pt] (O) circle (1);
    \node[below, xshift=-8pt] at (Px)
    {\footnotesize $\sqrt{\bar{\alpha}_k}$};
    \node[xshift=-6pt, rotate=90] at (Py)
    {\footnotesize $\sqrt{\,1-\bar{\alpha}_k\,}$};

    \fill [blue] (P) circle (0.03);
    \draw[densely dashed] (P) -- (Px);
    \draw[densely dashed] (P) -- (Py);
    \fill (Px) circle (0.02);
    \fill (Py) circle (0.02);
    \fill (X1) circle (0.03);
    \fill (Y1) circle (0.03);

    \node[above right] at (Y1) {$k=K$};
    \node[above right] at (X1) {$k=0$};
    
    \node[right, yshift=2pt] at (P) {$\mathbf{z}_k = \sqrt{\bar{\alpha}_k} \mathbf{z}_0 + \sqrt{1 - \bar{\alpha}_k}\, \boldsymbol{\epsilon}$};
    
    \node[right, xshift=-12pt, yshift=20pt] at (P) {$\omega_{k}'(k)\,\mathbf{v}_{k}$};
    
    \node[left,xshift=-20pt, yshift=2pt] at (Y1) {$\mathbf{v}_k =  - \sqrt{1 - \bar{\alpha}_k} \mathbf{z}_0 + \, \sqrt{\bar{\alpha}_k} \boldsymbol{\epsilon}$};

    \node[below right, xshift=5pt, yshift=-3pt] at (X1) {$\mathbf{z}_0$};
    \node[above left, xshift=0pt, yshift=8pt] at (Y1) {$\boldsymbol{\epsilon}$};

    \draw[thick] (O) -- (P);

    \pic[draw,-,angle radius=12pt,angle eccentricity=1.25, "$\omega_k$"{font=\scriptsize, inner sep=0.5pt, xshift=5pt, yshift=1pt}] {angle = X1--O--P};

    \draw[-{Triangle[length=1.5mm,width=1.2mm]}, thick, black]($(O)!0!(O)$) -- ++({cos(\om)},{sin(\om)});
    \draw[-{Triangle[length=1.5mm,width=1.2mm]}, thick, black]($(O)!0!(O)$) -- ++({cos(\om + 90)},{sin(\om + 90)});
    \draw ($(O)+({0.1*cos(\om)},{0.1*sin(\om)})$)
      -- ++({0.1*cos(\om+90)},{0.1*sin(\om+90)})
      -- ++({-0.1*cos(\om)},{-0.1*sin(\om)});
    
    \draw[-{Triangle[length=2mm,width=1.5mm]}, ultra thick, blue]
    ($(P)!0!(O)$) ++({-sin(\om)*0.02},{cos(\om)*0.02})
    -- ++({-sin(\om)*0.4},{cos(\om)*0.4});

\end{tikzpicture}
\caption{%
Geometric illustration of the $v$-prediction formulation on the unit circle.
}
\label{fig:v_prediction}
\end{figure*}
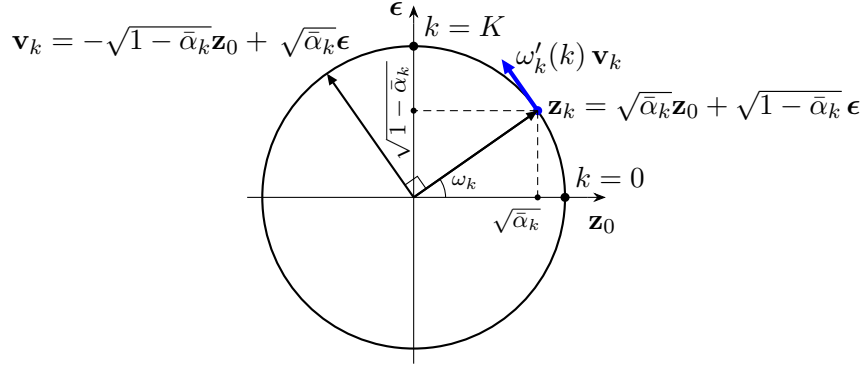

Because $\mathbf{z}_0$ is a member in a Gaussian latent space, it follows that each $\mathbf{z}_k$ is itself Gaussian for all $k$ with mean $\mathbf{0}$ and constant unit variance $\mathbf{I}$:

\begin{equation}
\mathbb{E}[\mathbf{z}_k] = \sqrt{\bar{\alpha}_k}\,\mathbb{E}[\mathbf{z}_0] + \sqrt{1-\bar{\alpha}_k}\,\mathbb{E}[\boldsymbol{\epsilon}] = \sqrt{\bar{\alpha}_{k}}\,\mathbf{0} + \sqrt{1-\bar{\alpha}_{k}}\,\mathbf{0} = \mathbf{0}, 
\end{equation}

\begin{equation}
\mathrm{Var}(\mathbf{z}_k) = \bar{\alpha}_k\,\mathrm{Var}(\mathbf{z}_0) + (1-\bar{\alpha}_k)\,\mathrm{Var}(\boldsymbol{\epsilon}) = \bar{\alpha}_k\,\mathbf{I} + (1-\bar{\alpha}_k)\,\mathbf{I} = \mathbf{I}.
\end{equation}

Hence, the diffusion process can be viewed as a rotation on the unit $(\mathbf{z}_0,\boldsymbol{\epsilon})$ plane. By introducing an angular parameter $\omega_k$ such that $\sqrt{\bar{\alpha}_k}=\cos\omega_k$ and $\sqrt{1-\bar{\alpha}_k}=\sin\omega_k$, the sample $\mathbf{z}_k$ can be expressed as

\begin{equation}
    \mathbf{z}_k = \cos\omega_k\,\mathbf{z}_0 + \sin\omega_k\,\boldsymbol{\epsilon},
\end{equation}

which parameterizes a circular trajectory from pure data ($\omega_0=0$) to pure noise ($\omega_K=\tfrac{\pi}{2}$). Differentiating this expression with respect to $\omega_k$ yields

\begin{equation}
\label{eq:velocity_definition}
\frac{\partial \mathbf{z}_k}{\partial \theta_k} = -\sin\omega_k\,\mathbf{z}_0 + \cos\omega_k\,\boldsymbol{\epsilon} = - \sqrt{1-\bar{\alpha}_k}\,\mathbf{z}_0 + \sqrt{\bar{\alpha}_k}\,\boldsymbol{\epsilon} \equiv \mathbf{v}_k.
\end{equation}

Because $\mathbf{z}_0$ and $\boldsymbol{\epsilon}$ are orthogonal and have unit variance, $\mathbf{z}_k$ and $\mathbf{v}_k$ also form an orthonormal pair. Geometrically, $\mathbf{z}_k$ represents the position on the unit circle (or hypersphere) and the vector $\mathbf{v}_k$ defines the tangent (velocity) direction of the probability-flow trajectory. The derivative with respect to $k$ shows us that the magnitude of the diffusion motion is modulated by $\omega_{k}'(k)$ and the direction is fully determined by $\mathbf{v}_k$:

\begin{equation}
\label{eq:derivative_wrt_t}
\frac{d\mathbf{z}_{k}}{dk} = \frac{d\omega_{k}}{dk}\,\frac{\partial \mathbf{x}_{k}}{\partial\omega_{k}} = \omega_{t}'(k)\,\mathbf{v}_{k},
\end{equation}.

\begin{algorithm}[t]
\caption{SimCast-S2S Training}
\label{algo:simcast_training}
\begin{algorithmic}[0]
\Require Dataset $\mathbb{D}$
\Require Pretrained VAE encoders $\mathcal{E}_g$ for $g \in \{\mathrm{wind}, \mathrm{mass}, \mathrm{thermal}, \mathrm{hydro}, \mathrm{precip}\}$
\Require Neural denoiser $f_{\boldsymbol{\theta}}$
\Require Diffusion schedule $\{\bar{\alpha}_k\}_{k=1}^{K}$, with $\bar{\alpha}_0 = 1$

\Repeat
    \State Sample
    $\left(\mathbf{x}_{\mathrm{wind}}, \mathbf{x}_{\mathrm{mass}}, \mathbf{x}_{\mathrm{thermal}}, \mathbf{x}_{\mathrm{hydro}}, \mathbf{x}_{\mathrm{precip}}, t, \mathbf{y}_{\mathrm{precip}}\right) \in \mathbb{D}$
    \For{$g \in \{\mathrm{wind}, \mathrm{mass}, \mathrm{thermal}, \mathrm{hydro}, \mathrm{precip}\}$}
        \State Compute: $\left(\boldsymbol{\mu}_g, \boldsymbol{\sigma}_g^2\right)= \mathcal{E}_g \left(\mathbf{x}_g\right)$
        \State Sample: $\boldsymbol{\epsilon}_g \sim \mathcal{N}(\mathbf{0}, \mathbf{I})$
        \State Compute conditioning latent: $\mathbf{z}_g = \boldsymbol{\mu}_g + \boldsymbol{\sigma}_g \odot \boldsymbol{\epsilon}_g$
    \EndFor
    \State Concatenate:
    $\mathbf{Z}_{\mathrm{cond}} = \mathbf{z}_{\mathrm{wind}} \oplus \mathbf{z}_{\mathrm{mass}} \oplus \mathbf{z}_{\mathrm{thermal}} \oplus \mathbf{z}_{\mathrm{hydro}} \oplus \mathbf{z}_{\mathrm{precip}}.$
    \State Compute: $\left(\boldsymbol{\mu}_{y},\boldsymbol{\sigma}_{y}^{2}\right) = \mathcal{E}_{\mathrm{precip}} \left(\mathbf{y}_{\mathrm{precip}}\right)$
    \State Sample:
    $\boldsymbol{\epsilon}_{y} \sim \mathcal{N}(\mathbf{0}, \mathbf{I})$
    \State Compute clean latent:
    $\mathbf{z}_0 = \boldsymbol{\mu}_{y} + \boldsymbol{\sigma}_{y} \odot \boldsymbol{\epsilon}_{y}$
    \State Sample:
    $k \sim \mathrm{Uniform}\left(\{1,\ldots,K\}\right)$
    \State Sample:
    $\boldsymbol{\epsilon} \sim \mathcal{N}(\mathbf{0}, \mathbf{I})$
    \State Compute noisy latent:
    $\mathbf{z}_k = \sqrt{\bar{\alpha}_k}\,\mathbf{z}_0 + \sqrt{1-\bar{\alpha}_k}\,\boldsymbol{\epsilon}.$
    \State Compute true velocity: $\mathbf{v}_{k} = -\sqrt{1-\bar{\alpha}_{k}}\,\mathbf{z}_{0} + \sqrt{\bar{\alpha}_{k}}\,\boldsymbol{\epsilon}$.
    \State Predict velocity: $\hat{\mathbf{v}}_k = f_{\boldsymbol{\theta}}\left(\mathbf{Z}_{\mathrm{cond}}, t, \mathbf{z}_k, k\right)$
    \State Take a gradient step on: $\nabla_{\boldsymbol{\theta}}\left\|\mathbf{v}_k - \hat{\mathbf{v}}_k \right\|_2^2.$
\Until{converged}
\end{algorithmic}
\end{algorithm}

Since $\mathbf{v}_k$ is still a unit vector, i.e., $\mathrm{Var}(\mathbf{v}_k) = \mathbf{I}$, the transition from the $\boldsymbol{\epsilon}$-prediction to the $v$-prediction formulation can be interpreted as a change of coordinate basis from the original $(\mathbf{z}_0,\boldsymbol{\epsilon})$ system to the rotated $(\mathbf{z}_k,\mathbf{v}_k)$ system. As shown in Fig. \ref{fig:v_prediction}, this transformation corresponds to a $90^\circ$ counterclockwise rotation on the unit circle in the $(\mathbf{z}_0,\boldsymbol{\epsilon})$ plane, preserving orthogonality and unit norm. Consequently,  conversion between two basis becomes immediately straightforward:

\begin{subequations}
\begin{minipage}{0.45\textwidth}
\begin{equation}
\label{eq:conversion_a}
\left\{
\begin{aligned}
\mathbf{z}_0 &= \sqrt{\mathbf{\bar{\alpha}}_k}\,\mathbf{z}_k - \sqrt{1 - \mathbf{\bar{\alpha}}_k}\,\mathbf{v}_k,\\[2pt]
\boldsymbol{\epsilon} &= \sqrt{1 - \mathbf{\bar{\alpha}}_k}\,\mathbf{z}_k + \sqrt{\mathbf{\bar{\alpha}}_k}\,\mathbf{v}_k
\end{aligned}
\right.
\tag{\theparentequation a}
\end{equation}
\end{minipage}\hfill
\begin{minipage}{0.45\textwidth}
\begin{equation}
\label{eq:conversion_b}
\left\{
\begin{aligned}
\mathbf{z}_k &= \sqrt{\mathbf{\bar{\alpha}}_k}\,\mathbf{z}_0 + \sqrt{1 - \mathbf{\bar{\alpha}}_k}\,\boldsymbol{\epsilon},\\[2pt]
\mathbf{v}_k &= -\sqrt{1 - \mathbf{\bar{\alpha}}_k}\,\mathbf{z}_0 + \sqrt{\mathbf{\bar{\alpha}}_k}\,\boldsymbol{\epsilon}
\end{aligned}
\right.
\tag{\theparentequation b}
\end{equation}
\end{minipage}
\end{subequations}

The signal-to-noise ratio (SNR) at timestep $k$ is defined as the ratio between the signal variance and the noise variance:
\begin{equation}
\mathrm{SNR}_k = \frac{\mathrm{Var}(\sqrt{\bar{\alpha}_k}\,\mathbf{z}_0)}{\mathrm{Var}(\sqrt{1-\bar{\alpha}_k}\,\boldsymbol{\epsilon})} = \frac{\bar{\alpha}_k}{1-\bar{\alpha}_k}.
\end{equation}

The standard $\boldsymbol{\epsilon}$-prediction objective minimizes the MSE between the true and predicted noise: $\mathcal{L}_{\boldsymbol{\epsilon}} = \mathbb{E}_k\,\|\boldsymbol{\epsilon} - \boldsymbol{\hat{\epsilon}}\|^2$. To examine how this loss contributes to the construction loss $\mathcal{L}_{\mathbf{z}_0} = \mathbb{E}_k\,\|\mathbf{z}_0 - \hat{\mathbf{z}}_0\|^2$ at different denoising step $k$, we rewrite $\mathcal{L}_{\mathbf{z}_0}$ in terms of the true clean sample $\mathbf{z}_0 = \frac{1}{\sqrt{\bar{\alpha}_k}}[\mathbf{z}_k - \sqrt{1-\bar{\alpha}_k}\,\boldsymbol{\epsilon}]$ and the predicted clean sample $\hat{\mathbf{z}}_0 = \frac{1}{\sqrt{\bar{\alpha}_k}}[\mathbf{z}_k - \sqrt{1-\bar{\alpha}_k}\,\boldsymbol{\hat{\epsilon}}]$. The reconstruction loss is then:

\begin{equation}
\mathcal{L}_{\mathbf{z}_0} = \mathbb{E}_k \big\|\mathbf{z}_0 - \hat{\mathbf{z}}_0\big\|^2 = \mathbb{E}_k \left[ \frac{1-\bar{\alpha}_k}{\bar{\alpha}_k}\big\|\boldsymbol{\epsilon} - \boldsymbol{\hat{\epsilon}}\big\|^2 \right] = \mathbb{E}_k \left[ \frac{1}{\text{SNR}_k}\,\big\|\boldsymbol{\epsilon} - \boldsymbol{\hat{\epsilon}}\big\|^2 \right].
\end{equation}

We see that each term in $\mathcal{L}_{\boldsymbol{\epsilon}}$ contributes differently to the target reconstruction loss $\mathcal{L}_{\mathbf{z}_0}$ by $\frac{1}{\text{SNR}_k}$. This weight blows up as $\bar{\alpha}_k \to 0$ (i.e. $k \to K$). Although $\boldsymbol{\epsilon}$-prediction still provides an unbiased estimator of the score function, its inverse-SNR imbalance implies poor numerical conditioning and motivates the $v$-prediction formulation, whose contribution remains stable over time.

Indeed, writing the reconstruction loss $\mathcal{L}_{\mathbf{z}_0}$ with $\mathbf{z}_0 = \sqrt{\mathbf{\bar{\alpha}}_k}\,\mathbf{z}_k - \sqrt{1 - \mathbf{\bar{\alpha}}_k}\,\mathbf{v}_k$ and $\hat{\mathbf{z}}_0 = \sqrt{\mathbf{\bar{\alpha}}_k}\,\mathbf{z}_k - \sqrt{1 - \mathbf{\bar{\alpha}}_k}\,\hat{\mathbf{v}}_k$, it now becomes:

\begin{equation}
\mathcal{L}_{\mathbf{z}_0} = \mathbb{E}_k \big\|\mathbf{z}_0 - \hat{\mathbf{z}}_0\big\|^2 = \mathbb{E}_k \left[ (1-\bar{\alpha}_k)\,\big\|\boldsymbol{v}_k - \hat{\boldsymbol{v}}_k\big\|^2 \right].
\end{equation}

Therefore, minimizing $\mathcal{L}_{\mathbf{v}} = \mathbb{E}_k \big\|\boldsymbol{v}_k - \hat{\boldsymbol{v}}_\theta(\mathbf{z}_k, k)\big\|^2$ induces $\mathbf{z}_0$-errors that scale only by the bounded factors $1 - \bar{\alpha}_k \in \left[0, 1\right]$. As a result, although $v$-prediction optimizes a step-stationary target (unit variance for all $k$) just like $\epsilon$-prediction, it has no inverse-SNR amplification. The per-step gradient magnitudes remain commensurate across $k$, which implies much more stable optimization.

From the five pretrained VAEs corresponding to the five groups of physical fields listed in Table~\ref{tab:variable_groups}, together with the loss function in Eq.~\ref{eq:ddpm_loss}, the general sampling rule in Eq.~\ref{eq:ddim_interpolation}, and the basis conversions in Eq.~\ref{eq:conversion_a}--\ref{eq:conversion_b}, we obtain the complete training and generating algorithms for SimCast-S2S, summarized in Algorithms~\ref{algo:simcast_training}--\ref{algo:simcast_sampling}, respectively.

\begin{algorithm}[t]
\caption{SimCast-S2S Generating}
\label{algo:simcast_sampling}
\begin{algorithmic}[0]
\Require Input fields: $\mathbf{x}_{\mathrm{wind}}, \mathbf{x}_{\mathrm{mass}}, \mathbf{x}_{\mathrm{thermal}}, \mathbf{x}_{\mathrm{hydro}}, \mathbf{x}_{\mathrm{precip}}$
\Require Day-of-year information $t$
\Require Pretrained VAE encoders $\mathcal{E}_g$ for $g \in \{\mathrm{wind}, \mathrm{mass}, \mathrm{thermal}, \mathrm{hydro}, \mathrm{precip}\}$
\Require Pretrained precipitation decoder $\mathcal{D}_{\mathrm{precip}}$
\Require Trained neural denoiser $f_{\boldsymbol{\theta}}$
\Require Diffusion schedule $\{\bar{\alpha}_k\}_{k=0}^{K}$, with $\bar{\alpha}_0 = 1$
\Require Stochasticity parameter $\eta \in [0,1]$
\For{$g \in \{\mathrm{wind}, \mathrm{mass}, \mathrm{thermal}, \mathrm{hydro}, \mathrm{precip}\}$}
    \State Compute: $\left(\boldsymbol{\mu}_g, \boldsymbol{\sigma}_g^2\right)= \mathcal{E}_g \left(\mathbf{x}_g\right)$
    \State Sample: $\boldsymbol{\epsilon}_g \sim \mathcal{N}(\mathbf{0}, \mathbf{I})$
    \State Compute conditioning latent: $\mathbf{z}_g = \boldsymbol{\mu}_g + \boldsymbol{\sigma}_g \odot \boldsymbol{\epsilon}_g$
\EndFor
\State Concatenate:
$\mathbf{Z}_{\mathrm{cond}} = \mathbf{z}_{\mathrm{wind}} \oplus \mathbf{z}_{\mathrm{mass}} \oplus \mathbf{z}_{\mathrm{thermal}} \oplus \mathbf{z}_{\mathrm{hydro}} \oplus \mathbf{z}_{\mathrm{precip}}.$
\State Initialize: $\hat{\mathbf{z}}_K\sim\mathcal{N}(\mathbf{0},\mathbf{I})$.
\For{$k=K,K-1,\ldots,1$}
    \State Predict velocity: $\hat{\mathbf{v}}_k=f_{\boldsymbol{\theta}}\left(\mathbf{Z}_{\mathrm{cond}},t,\hat{\mathbf{z}}_k,k\right)$.
    \State Compute clean latent estimate: $\hat{\mathbf{z}}_0=\sqrt{\bar{\alpha}_k}\,\hat{\mathbf{z}}_k-\sqrt{1-\bar{\alpha}_k}\,\hat{\mathbf{v}}_k$.
    \State Compute noise estimate: $\hat{\boldsymbol{\epsilon}}_k=\sqrt{1-\bar{\alpha}_k}\,\hat{\mathbf{z}}_k+\sqrt{\bar{\alpha}_k}\,\hat{\mathbf{v}}_k$.
    \State Compute: $\sigma_k=\eta\sqrt{\frac{1-\bar{\alpha}_{k-1}}{1-\bar{\alpha}_k}}\sqrt{1-\frac{\bar{\alpha}_k}{\bar{\alpha}_{k-1}}}$.
    \State Compute: $c_k=\sqrt{1-\bar{\alpha}_{k-1}-\sigma_k^2}$.
    \State Sample: $\boldsymbol{\zeta}_k\sim\mathcal{N}(\mathbf{0},\mathbf{I})$
    \State Compute: $\hat{\mathbf{z}}_{k-1}=\sqrt{\bar{\alpha}_{k-1}}\,\hat{\mathbf{z}}_0+c_k\hat{\boldsymbol{\epsilon}}_k+\sigma_k\boldsymbol{\zeta}_k$.
\EndFor
\State \Return precipitation-anomaly forecast: $\hat{\mathbf{y}}=\mathcal{D}_{\mathrm{precip}}\left(\hat{\mathbf{z}}_0\right)$.
\end{algorithmic}
\end{algorithm}

In this study, $\mathbf{Z}_{\mathrm{cond}}$ and $t$ encode the physical atmospheric states over the previous $28$ days and their associated seasonal context, respectively. Both $\mathbf{Z}_{\mathrm{cond}}$ and $t$ are step-independent and remain fixed throughout the denoising process. The neural network uses these inputs to guide the reverse process so that the generated latent $\hat{\mathbf{z}}_0$ represents the precipitation anomaly over lead days 15--28, corresponding to the 14-day subseasonal forecast window. A conceptual illustration of the latent diffusion process is provided in Fig.~\ref{fig:diffusion}. During generation, the neural network denoises the latent state sequentially from step $K$ to step $0$. It should be noted that the generation process is initialized randomly and $\hat{\mathbf{z}}_K$ can be far apart from $\mathbf{z}_K$. However, the neural network uses the conditioning information to guide the denoising trajectory so that $\hat{\mathbf{z}}_0$ is expected to be close to $\mathbf{z}_0$. Additionally, if $\eta = 0$, the denoising trajectory (yellow) is deterministic for a fixed initial latent $\hat{\mathbf{z}}_K$. If $0 < \eta \le 1$, additional Gaussian noise is injected at each reverse step, which makes the denoising trajectory stochastic.

The neural network $f_{\theta}$ is designed as a compact denoising engine for latent-space forecasting. It follows an encoder--decoder structure with multiple downsampling blocks, each combining convolutional and transformer layers, followed by a symmetric set of upsampling blocks. The convolutional layers refine local spatial structure within the latent fields and the transformer layers handle the temporal exchange of information between the conditioning history and the forecast target. 

This temporal structure is explicit. The conditioning sequence contains 28 tokens, one for each input day, and the target sequence contains 14 tokens, one for each forecast day. Through cross-attention~\citep{vaswani2017attention}, the 14 noisy target tokens query the 28 conditioning tokens and extract the parts of the past atmospheric evolution most relevant for denoising the future precipitation latent. In this formulation, $\mathbf{Z}_{\mathrm{cond}}$ provides the key--value pairs, and the intermediate representation of $\hat{\mathbf{z}}_k$ provides the queries. The denoiser therefore selectively learns which parts of the 28-day atmospheric history should inform each forecast day.

\begin{figure}
    \centering
    \includegraphics[width=1\linewidth]{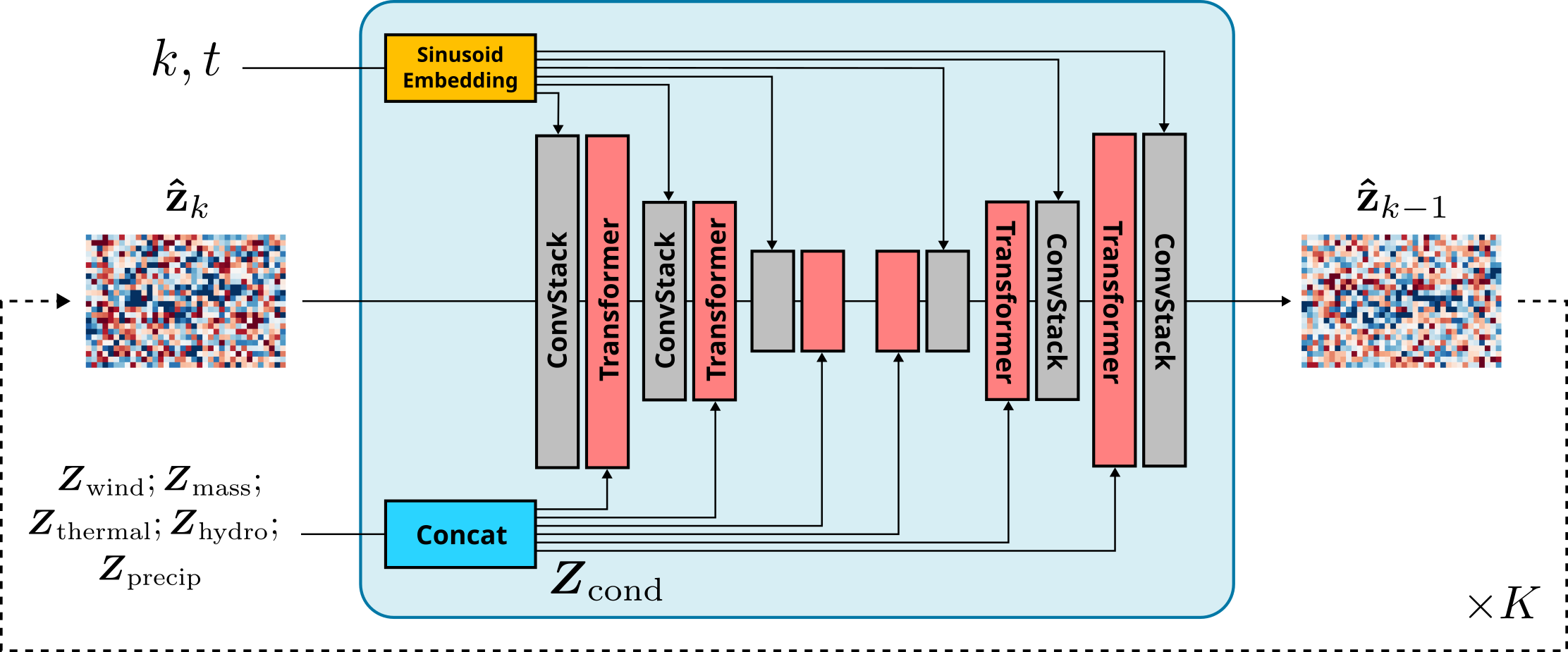}
    \caption{
    Architecture of the SimCast-S2S denoiser. At step $k$, the noisy target precipitation latent $\hat{\mathbf{z}}_k$ is passed through a sequence of ConvStack and Transformer blocks to predict the velocity $\hat{\mathbf{v}}_k$. Although the denoiser is trained to predict $\hat{\mathbf{v}}_k$, the next denoised latent state $\hat{\mathbf{z}}_{k-1}$ is obtained using Eqs.~\ref{eq:conversion_a}--\ref{eq:conversion_b} and Eq.~\ref{eq:ddim_interpolation}. The conditioning information is formed by concatenating the latent representations of the five physical groups, $\mathbf{z}_{\mathrm{wind}}$, $\mathbf{z}_{\mathrm{mass}}$, $\mathbf{z}_{\mathrm{thermal}}$, $\mathbf{z}_{\mathrm{hydro}}$, and $\mathbf{z}_{\mathrm{precip}}$, into $\mathbf{Z}_{\mathrm{cond}}$, which is injected into the Transformer blocks through cross-attention. The diffusion step $k$ and seasonal information $t$ are encoded using separate sinusoidal embedding layers and injected into the ConvStack blocks. This denoising operation is repeated for $K$ reverse steps, transforming an initial Gaussian latent sample $\hat{\mathbf{z}}_K \sim \mathcal{N}(\mathbf{0},\mathbf{I})$ into the generated target precipitation latent $\hat{\mathbf{z}}_0$.
    }
\label{fig:denoiser}
\end{figure}

The diffusion step $k$ and day-of-year $t$ are passed through separate sinusoidal embedding blocks \cite{vaswani2017attention}, which convert scalar inputs into vector representations while preserving their relative proximity in Euclidean space. Given the noisy target latent $\hat{\mathbf{z}}k$, the conditioning latent sequence $\mathbf{Z}_{\mathrm{cond}}$, the diffusion-step embedding, and the day-of-year embedding, the denoiser predicts the velocity $\hat{\mathbf{v}}_k$. This predicted velocity defines the direction used to move the latent state from $\hat{\mathbf{z}}_k$ toward $\hat{\mathbf{z}}_{k-1}$ at each reverse step. The detailed architecture of $f_{\theta}$ is provided in Fig.~\ref{fig:denoiser}

\subsection{Transfer Learning: LoRA}
\label{subsec:lora}

Perhaps one of the most important features of SimCast-S2S is its computationally efficient latent-space framework. This efficiency allows SimCast-S2S to be pretrained on 28 CESM2 atmospheric simulations and then transferred to real-world atmospheric estimates from ERA5. Traditional transfer learning typically follows one of common strategies: fine-tuning the full pretrained network with a small learning rate~\cite{pan2010survey, yosinski2014transferable, zhuang2021comprehensive}, freezing early layers and retraining only task-specific layers~\cite{donahue2014decaf, oquab2014learning}, or keeping the pretrained backbone mostly fixed while adding small trainable adaptation modules~\cite{rebuffi2017learning, houlsby2019parameter}. SimCast-S2S follows the third strategy by using low-rank adaptation (LoRA)~\cite{hu2022lora}. 

LoRA transfer-learning strategy is applied to $\mathcal{E}_{\mathrm{precip}}$, $\mathcal{D}_{\mathrm{precip}}$, and $f_{\theta}$. We do not adapt the VAE encoders and decoders for group wind, mass, thermal, hydro because these VAEs show no substantial degradation when evaluated on ERA5 with no fine-tuning. In contrast, precipitation exhibits a stronger distributional shift between CESM2 and ERA5, so the precipitation encoder and decoder are fine-tuned together with the denoising network. 

Each ConvStack layer in Figs.~\ref{fig:vae_architecture} and~\ref{fig:denoiser} contains multiple convolutional layers~\cite{lecun1998gradient, krizhevsky2012imagenet}, and each Transformer layer in Fig.~\ref{fig:denoiser} also contains multiple dense layers as their core components~\cite{vaswani2017attention}. For any such layer, let $\mathbf{W} \in \mathbb{R}^{d_{\mathrm{in}} \times d_{\mathrm{out}}}$ denote the pretrained weight matrix learned from CESM2, where $d_{\mathrm{in}}$ and $d_{\mathrm{out}}$ are the input and output dimensions, respectively. 

Instead of fine-tuning the full weight matrix $\mathbf{W}$, which can be computationally expensive and may overwrite useful structure learned during CESM2 pretraining, LoRA freezes $\mathbf{W}$ and learns only a low-rank update $\Delta \mathbf{W}$:

\begin{equation}
\mathbf{W}' = \mathbf{W} + \Delta \mathbf{W}.
\end{equation}

The matrix $\Delta \mathbf{W}$ can be written in a low-rank factorization:

\begin{equation}
\Delta \mathbf{W} = \mathbf{B}\mathbf{A},
\end{equation}

where $\mathbf{B} \in \mathbb{R}^{d_{\mathrm{in}} \times r}$, $\mathbf{A} \in \mathbb{R}^{r \times d_{\mathrm{out}}}$, and most importantly, $r \ll \min(d_{\mathrm{in}}, d_{\mathrm{out}})$ is the rank of matrix $\Delta \mathbf{W}$. For an input feature vector $\mathbf{h} \in \mathbb{R}^{d_\mathrm{in}}$, the adapted output becomes

\begin{equation}
\mathbf{h}\mathbf{W}' = \mathbf{h}\mathbf{W} + \mathbf{h}\mathbf{B}\mathbf{A}.
\end{equation}
The first term on the right-hand side preserves the pretrained CESM2 knowledge. The second term introduces a small trainable correction that adapts the model from CESM2 to ERA5. During fine-tuning, only $\mathbf{A}$ and $\mathbf{B}$ are optimized, $\mathbf{W}$ remains intact. As a result, we only need to optimize $\mathcal{O}(r(d_\mathrm{in} + d_\mathrm{out}))$ number of parameters, instead of $\mathcal{O}(d_\mathrm{in} d_\mathrm{out})$. Because $r$ is chosen to be much smaller than both $d_{\mathrm{in}}$ and $d_{\mathrm{out}}$, $\mathcal{O}(r(d_\mathrm{in} + d_\mathrm{out}))$ is orders of magnitude smaller than $\mathcal{O}(d_\mathrm{in} d_\mathrm{out})$. This makes LoRA more memory efficient and less prone to overfitting.

LoRA is particularly useful for SimCast-S2S because the model can exploit a large ensemble of CESM2 simulations. As shown in Section~\ref{subsec:eval}, CESM2 pretraining followed by ERA5 fine-tuning with LoRA performs substantially better than training on ERA5 alone, and it is one of the main reasons SimCast-S2S outperforms the process-based ECMWF-S2S. The only task assigned to SimCast-S2S during fine-tuning is to adjust for the domain shift from many simulated worlds to one single real world.

\subsection{Deep Learning Baselines: CNN, UNet}

SimCast-S2S is benchmarked against two widely used deep learning architectures for spatial prediction: CNN and UNet. CNN models are built on top of the convolution operator and provide a natural baseline for gridded prediction tasks and provide a natural baseline for gridded prediction tasks because they learn local spatial patterns through shared kernels~\citep{lecun1998gradient, krizhevsky2012imagenet}. The UNet architecture extends this idea with an encoder--decoder structure and skip connections that help preserve multiscale spatial information~\citep{ronneberger2015unet}. In this study, we construct three CNN baselines and three U-Net baselines with different model sizes. Tables~\ref{tab:cnn_baselines}--\ref{tab:unet_baselines} summarize the key design choices.

\begin{table}[t]
\centering
\caption{
Architectural configurations of the CNN baselines. Each CNN uses a stack of convolutional layers ($N$) with a fixed embedding dimension ($d$). The three model sizes increase both the embedding dimension and the number of layers.
}
\label{tab:cnn_baselines}
\small
\setlength{\tabcolsep}{15pt}
\renewcommand{\arraystretch}{1.15}
\begin{tabular}{lcc}
\toprule
\textbf{Model}
& $d$
& $N$ \\
\midrule
cnn-small
& $256$
& $4$ \\
cnn-medium
& $512$
& $8$ \\
cnn-large
& $1024$
& $12$ \\
\bottomrule
\end{tabular}
\end{table}

\begin{table}[t]
\centering
\caption{
Architectural configurations of the UNet baselines. Each U-Net uses a symmetric encoder--decoder structure with four sampling blocks, with embedding dimensions $\{d_i\}_{i=1}^{4}$, respectively. In the encoder, each block downsamples the spatial resolution and doubles the embedding dimension. In the decoder, the corresponding upsampling blocks recover the spatial resolution and halve the embedding dimension.
}
\label{tab:unet_baselines}
\small
\setlength{\tabcolsep}{8pt}
\renewcommand{\arraystretch}{1.15}
\begin{tabular}{lcccc}
\toprule
\textbf{Model}
& $d_1$
& $d_2$
& $d_3$
& $d_4$ \\
\midrule
unet-small
& $128$
& $256$
& $512$
& $1024$ \\
unet-medium
& $256$
& $512$
& $1024$
& $2048$ \\
unet-large
& $512$
& $1024$
& $2048$
& $4096$ \\
\bottomrule
\end{tabular}
\end{table}

\subsection{Physical Baseline: ECMWF-S2S}

To place SimCast-S2S against a strong process-based reference, we use ECMWF-S2S as the operational physical baseline. ECMWF is widely regarded as one of the world-leading numerical weather prediction centers, and its subseasonal forecasts, ECMWF-S2S, represent a highly developed dynamical forecasting system~\cite{bauer2015quiet, vitart2017s2s, vitart2022extended}. Unlike neural-network baselines, ECMWF-S2S is produced by integrating a physically based Earth-system model that explicitly resolves atmospheric dynamics, parameterizes unresolved physical processes, and generates ensemble forecasts through perturbed initial conditions and model uncertainties~\cite{leutbecher2008ensemble, vitart2017s2s}. Outperforming ECMWF-S2S would indicate that SimCast-S2S is competitive with one of the most advanced operational forecasting systems currently available. At the same time, this comparison should be interpreted with care. SimCast-S2S is not intended to replace ECMWF-S2S. Instead, it provides an inexpensive alternative for subseasonal precipitation forecasting, accessible to research groups and computing centers with smaller-scale computational resources.

\subsection{Evaluation Metrics}
\label{subsec:eval_method}

\subsubsection{Deterministic Skill Metrics}

We evaluate deterministic skill over all test samples using the ensemble-mean precipitation-anomaly forecast. Let $\hat{\mathbf{y}}^{(m)}$ and $\mathbf{y}^{(m)}$ denote the predicted and verifying precipitation-anomaly field for test sample $m$, respectively. Here, $m=1,\ldots,M$ indexes the test samples and $i=1,\ldots,N$ indexes the spatial grid points.

The first deterministic metric is mean absolute error (MAE), which measures the average magnitude of the grid-level forecast error across samples and space:

\begin{equation}
\mathrm{MAE} = \frac{1}{MN}\sum_{m=1}^{M}\sum_{i=1}^{N}\left|\hat{y}^{(m)}_i - y^{(m)}_i\right|  \quad \in \mathbb{R}.
\end{equation}

Lower MAE indicates smaller absolute error and therefore better deterministic accuracy. Because all precipitation fields are converted to standardized anomalies, MAE measures error relative to the local climatology.

The second deterministic metric is anomaly correlation coefficient (ACC), which evaluates whether the forecast captures the spatial pattern of the observed precipitation anomaly. For each test sample, ACC is first computed over the spatial grid:

\begin{equation}
\mathrm{ACC}^{(m)} = \frac{\sum_{i=1}^{N}\hat{y}^{(m)}_i y^{(m)}_i}{\sqrt{\sum_{i=1}^{N}\left(\hat{y}^{(m)}_i\right)^2}\sqrt{\sum_{i=1}^{N}\left(y^{(m)}_i\right)^2}} \quad \in \mathbb{R}.
\end{equation}

The reported ACC is then averaged across all test samples:

\begin{equation}
\mathrm{ACC} = \frac{1}{M}\sum_{m=1}^{M}\mathrm{ACC}^{(m)} \quad \in \mathbb{R}.
\end{equation}

Higher ACC indicates stronger agreement between the predicted and observed anomaly patterns. A skillful subseasonal precipitation forecast should achieve both low MAE and high ACC, meaning that it is close to the observed anomaly field in magnitude and also reproducing the spatial structure of wet and dry anomalies.

\subsubsection{Probabilistic Skill Metrics}

Probabilistic skill is evaluated using the full ensemble distribution the ensemble mean. For each test sample $m$, SimCast-S2S produces an ensemble of precipitation-anomaly forecasts
$\{\hat{\mathbf{y}}^{(m,e)}\}_{e=1}^{E}$, where $E$ is the number of ensemble members. These ensemble members define an empirical predictive distribution at each grid point. The probabilistic metrics evaluate whether this distribution assigns high probability to the observed outcome and whether it improves over a reference forecast distribution.

The first probabilistic metric is the ranked probability skill score (RPSS), which is used for tercile-based categorical forecasts. For each grid point, the precipitation anomaly distribution is divided into three climatological categories: below-normal, near-normal, and above-normal. Let $\mathbf{p}^{(m)}=(p^{(m)}_1,p^{(m)}_2,p^{(m)}_3)$ denote the forecast probabilities assigned to the three categories for sample $m$, and let $\mathbf{o}^{(m)}=(o^{(m)}_1,o^{(m)}_2,o^{(m)}_3)$ denote the observed one-hot category vector. The ranked probability score (RPS) is

\begin{equation}
\mathrm{RPS}^{(m)} = \sum_{c=1}^{3}\left(\sum_{j=1}^{c}p^{(m)}_j - \sum_{j=1}^{c} o^{(m)}_j\right)^2  \quad \in \mathbb{R}^{N}.
\end{equation}

The RPSS compares the forecast RPS against the climatological reference. The sample-level RPSS is defined as

\begin{equation}
\mathrm{RPSS}^{(m)} = 1 - \frac{\overline{\mathrm{RPS}}^{(m)}_{\mathrm{forecast}}}{\overline{\mathrm{RPS}}^{(m)}_{\mathrm{climatology}}}  \quad \in \mathbb{R}.
\end{equation}

Here and throughout the rest of this section, the overbar denotes averaging over all $N$ grid points. The reported RPSS is obtained by averaging the sample-level skill scores over all $M$ test samples. Positive RPSS indicates that the probabilistic forecast improves upon climatology. Negative RPSS indicates worse performance than climatology.

A more general version of RPSS is the continuous ranked probability skill score (CRPSS), which evaluates the full continuous predictive distribution rather than discrete tercile categories. For a predictive cumulative distribution function $F^{(m)}$ and verifying observation $y^{(m)}$ of sample $m$, the continuous ranked probability score (CRPS) is

\begin{equation}
\mathrm{CRPS}\left(F^{(m)}, y^{(m)}\right) = \int_{-\infty}^{\infty}\left[F^{(m)}(x) - \mathbf{1}_{y^{(m)} \le x}\right]^2 dx \quad \in \mathbb{R}^N.
\end{equation}

For ensemble forecasts, $F^{(m)}$ is represented by the empirical distribution of ensemble members. The CRPSS of sample $m$ is then defined as

\begin{equation}
\mathrm{CRPSS}^{(m)} = 1 - \frac{\overline{\mathrm{CRPS}}^{(m)}_{\mathrm{forecast}}}{\overline{\mathrm{CRPS}}^{(m)}_{\mathrm{climatology}}}  \quad \in \mathbb{R}.
\end{equation}

Similar to RPSS, the reported RPSS is the average of all sample-level skill scores in the test dataset. Higher CRPSS is better, positive CRPSS indicates improvement from the climatology.

The third probabilistic metric is the Brier skill score (BSS), which is used to evaluate binary event probabilities. In this study, BSS is applied to extreme precipitation events exceeding the $90$th percentile. Let $p^{(m)}$ denote the forecast probability of the event estimated from the ensemble forecasts, and let the one-hot category $o^{(m)} \in \{0,1\}$ denote whether the extreme event occurred. The Brier score of sample $m$ is

\begin{equation}
\mathrm{BS}^{(m)} = \left(p^{(m)} - o^{(m)}\right)^2 \quad \in \mathbb{R}^N.
\end{equation}

The BSS also compares this score with the corresponding climatological event probability:

\begin{equation}
\mathrm{BSS}^{(m)} = 1 - \frac{\overline{\mathrm{BS}}^{(m)}_{\mathrm{forecast}}}{\overline{\mathrm{BS}}^{(m)}_{\mathrm{climatology}}} \quad \in \mathbb{R}.
\end{equation}

The reported BSS is also the average score of all test samples. Positive BSS indicates that the ensemble forecasts capture extreme events better than the climatological reference.

Although RPSS, CRPSS, and BSS evaluate different aspects of probabilistic forecast quality, they share a common idea. A skillful subseasonal precipitation forecast should place probability mass near the observed precipitation anomaly. The three metrics differ mainly in how they break down the forecast probability density function (PDF). RPSS evaluates probability mass over discretized partitions of the PDF, CRPSS evaluates the continuous PDF as it is, and BSS evaluates the right tail of the PDF for extreme precipitation events.

\subsubsection{Uncertainty Quantification}

A powerful metric for quantifying the uncertainty of ensemble forecasts is the probability integral transform (PIT), which directly tests whether the predictive distribution is statistically consistent with the observations. The idea is simple. For an unbiased probabilistic forecast, the observation should behave like a random draw from the forecast distribution. Therefore, the PIT values should be uniformly distributed on $[0,1]$.

For test sample $m$ and grid point $i$, let $\widehat{F}_{m,i}$ denote the empirical cumulative distribution function (CDF) defined by the $E$--member ensemble forecast $\{\hat{y}^{(m,e)}_i\}_{e=1}^{E}$. The PIT value is computed as
\begin{equation}
    u_{m,i} = \widehat{F}_{m,i}\left(y^{(m)}_i\right) = \frac{1}{E} \sum_{e=1}^{E}\mathbf{1}\left\{\hat{y}^{(m,e)}_i \le y^{(m)}_i\right\}.
\end{equation}
where $y^{(m)}_i$ is the true precipitation anomaly. An unbiased ensemble should produce PIT values that are approximately uniform. A $\cup$--shaped PIT histogram indicates underdispersion, meaning that the ensemble spread is too narrow and observations fall too often in the tails of the forecast distribution. A $\cap$--shaped PIT histogram indicates overdispersion, meaning that the ensemble spread is too wide. 

To quantify the deviation of the PIT histogram from uniformity, we compute the $\chi^2$--distance to flatness. Let the interval $[0,1]$ be divided into $B$ equally spaced bins, and let

\begin{equation}
    n_b = \sum_{m=1}^{M} \sum_{i=1}^{N} \mathbf{1}\left\{u_{m,i} \in I_b\right\}
\end{equation}

denote the number of PIT values falling in bin $I_b$. The total number of PIT values is $MN$, so the expected count under a perfectly uniform PIT distribution is $\frac{MN}{B}$ in each bin. The $\chi^2$ distance to flatness is then defined as
\begin{equation}
    \chi^2 = \sum_{b=1}^{B}
    \frac{\left(n_b - \frac{MN}{B}\right)^2}{\frac{MN}{B}}.
\end{equation}

A lower value of $\chi^2$--distance indicates a PIT histogram closer to uniformity and therefore better capture the true uncertainty. A large $\chi^2$--distance value indicates that the ensemble distribution is systematically inconsistent with the observed precipitation anomalies, either because the ensemble is underdispersed, or overdispersed.

A completely different view on uncertainty quantification is provided by the relationship between ensemble spread and deterministic error. The spread--RMSE relationship evaluates whether the ensemble spread is consistent with the actual error of the ensemble-mean forecast. We want larger forecast uncertainty to correspond to larger ensemble-mean error. Consequently, the average ensemble spread should be comparable to the average RMSE. 

For each test sample $m$ and grid point $i$, the ensemble-mean forecast over all $E$ members is
\begin{equation}
    \bar{y}^{(m)}_i = \frac{1}{E} \sum_{e=1}^{E} \hat{y}^{(m,e)}_i ,
\end{equation}
and the ensemble spread is
\begin{equation}
    s^{(m)}_i = \sqrt{\frac{1}{E-1}\sum_{e=1}^{E}\left(\hat{y}^{(,e)}_i - \bar{y}^{(m)}_i\right)^2} .
\end{equation}

The spatially averaged ensemble spread for sample $m$ is then defined as
\begin{equation}
    \mathrm{Spread}^{(m)}
    =
    \sqrt{
    \frac{1}{N}
    \sum_{i=1}^{N}
    \left(
        s^{(m)}_i
    \right)^2
    },
\end{equation}

and the corresponding RMSE of the ensemble mean is simply

\begin{equation}
    \mathrm{RMSE}^{(m)} = \sqrt{ \frac{1}{N}\sum_{i=1}^{N}\left(\bar{y}^{(m)}_i - y^{(m)}_i\right)^2} .
\end{equation}

For an ideally statistically consistent ensemble, it can be shown that the spread--RMSE diagram should approximately be the $1{:}1$ line~\cite{fortin2014spread}. If the spread is much smaller than the RMSE, the ensemble is underdispersed and does not represent enough forecast uncertainty. If the spread is much larger than the RMSE, the ensemble is overdispersed and produces an unnecessarily broad forecast distribution.

We also evaluate uncertainty using the mean interval score (MIS). For a central $(1-\alpha)$ prediction interval, let $l^{(m)}_i$ and $u^{(m)}_i$ denote the lower and upper ensemble quantiles at grid point $i$ for sample $m$. The interval score is defined as

\begin{equation}
    \mathrm{IS}^{(m)}_i(\alpha) = \left(u^{(m)}_i - l^{(m)}_i\right) + \frac{2}{\alpha} \left(l^{(m)}_i -y^{(m)}_i\right)\mathbf{1}\left\{y^{(m)}_i < l^{(m)}_i\right\} + \frac{2}{\alpha} \left(y^{(m)}_i - u^{(m)}_i\right) \mathbf{1} \left\{y^{(m)}_i > u^{(m)}_i\right\}.
\end{equation}

The MIS is obtained by averaging over all test samples and grid points:
\begin{equation}
    \mathrm{MIS}(\alpha)
    =
    \frac{1}{MN}
    \sum_{m=1}^{M}
    \sum_{i=1}^{N}
    \mathrm{IS}^{(m)}_i(\alpha).
\end{equation}

Lower MIS indicates better interval forecasts. The first term $(u^{(m)}_i - l^{(m)}_i)$ rewards sharp prediction intervals. The terms $\frac{2}{\alpha} (l^{(m)}_i -y^{(m)}_i)\,\mathbf{1}\{y^{(m)}_i < l^{(m)}_i\}$ and $\frac{2}{\alpha} (y^{(m)}_i - u^{(m)}_i) \mathbf{1} \{y^{(m)}_i > u^{(m)}_i\}$ penalize observations that fall outside the predicted interval. Therefore, MIS jointly evaluates sharpness and statistical consistency. A good ensemble should produce intervals that are narrow but still contain the observed precipitation anomaly.

\subsubsection{Realism Quantification: Spatial Auto-correlation}

Beyond deterministic accuracy and probabilistic skill, we also care about whether the generated precipitation-anomaly fields have realistic spatial structure. This is important because a forecast can obtain reasonable pointwise scores but smooths out the physical structures at the same time. A good way to quantify realism is comparing the spatial autocorrelation between the forecasts and the true observations. Spatial autocorrelation quantifies
the internal spatial texture of a precipitation field by measuring how strongly the field
correlates with a spatially shifted version of itself. 

For a precipitation-anomaly field $\mathbf{y}^{(m)}$, we compute directional autocorrelation by shifting the field by a spatial lag $\ell$ and correlating the original field with the shifted field over their overlapping grid points. Let $\mathcal{S}_{d,\ell}$ denote the shift operator in direction $d$ with lag $\ell$, where $d \in \{\mathrm{zonal}, \mathrm{meridional}, \mathrm{main\text{-}diagonal}, \mathrm{anti\text{-}diagonal}\}.$

For example, a zonal shift compares each grid point with another grid point $\ell$ pixels away in the longitudinal direction, a meridional shift compares grid points $\ell$ pixels apart in the latitudinal direction. The two diagonal shifts compare grid points displaced simultaneously in both spatial dimensions.

Let $\Omega_{d,\ell}$ denote the set of grid points for which both the original field and the shifted field are defined. The observed directional autocorrelation for test sample $m$, in direction $d$, and with lag $\ell$ is defined as
\begin{equation}
    \rho^{(m)}_{\mathrm{obs}}(d,\ell)
    =
    \frac{
        \sum_{i \in \Omega_{d,\ell}}
        \left(
            y^{(m)}_i
            -
            \bar{y}^{(m)}_{d,\ell}
        \right)
        \left(
            \mathcal{S}_{d,\ell} y^{(m)}_i
            -
            \overline{\mathcal{S}_{d,\ell} y}^{(m)}_{d,\ell}
        \right)
    }{
        \sqrt{
        \sum_{i \in \Omega_{d,\ell}}
        \left(
            y^{(m)}_i
            -
            \bar{y}^{(m)}_{d,\ell}
        \right)^2
        }
        \sqrt{
        \sum_{i \in \Omega_{d,\ell}}
        \left(
            \mathcal{S}_{d,\ell} y^{(m)}_i
            -
            \overline{\mathcal{S}_{d,\ell} y}^{(m)}_{d,\ell}
        \right)^2
        }
    } ,
\end{equation}

where $\bar{y}^{(m)}_{d,\ell}$ and $\overline{\mathcal{S}_{d,\ell} y}^{(m)}_{d,\ell}$ are the spatial means of the original and shifted fields over the overlapping domain $\Omega_{d,\ell}$. The same calculation is applied to each forecast ensemble member. For ensemble member $e$, the forecast spatial autocorrelation is $\rho^{(m,e)}_{\mathrm{forecast}}(d,\ell)$

We compute this quantity for multiple lags $\ell$ and for the four directions shown in Fig.~\ref{fig:realism}. The resulting distributions are summarized with boxplots across test samples and ensemble members.

A realistic precipitation generator should reproduce the observed decay of spatial autocorrelation with increasing lag. In other words, the forecast autocorrelation $\rho^{(m,e)}_{\mathrm{forecast}}(d,\ell)$ should follow the same lag-dependent structure as the observed autocorrelation $\rho^{(m)}_{\mathrm{obs}}(d,\ell)$ across directions $d$ and spatial lags $\ell$. If $\rho^{(m,e)}_{\mathrm{forecast}}(d,\ell)$ remains too high relative to $\rho^{(m)}_{\mathrm{obs}}(d,\ell)$ as $\ell$ increases, the generated fields are overly smooth. If $\rho^{(m,e)}_{\mathrm{forecast}}(d,\ell)$ decays too rapidly, the generated fields are too noisy. 

\bibliographystyle{unsrt}  
\bibliography{references}  

\end{document}